%% file: main.tex
\def\usearxivstyle{1}
\ifdefined\usearxivstyle

\documentclass[11pt]{article}
\usepackage[numbers]{natbib}
\usepackage{packages}
\usepackage{statistics}
\usepackage{statistics-macros}
\usepackage{manyfoot}

\definecolor{innerboxcolor}{rgb}{.9,.95,1}
\definecolor{outerlinecolor}{rgb}{.6,0,.2}
\definecolor{outerlinecoloreb}{rgb}{0,1,0}
\definecolor{innerboxcoloreb}{rgb}{1,1,1}

\newcommand{\dottedline}[1]{%
  \multicolumn{1}{c}{\dotfill} & \multicolumn{1}{c}{\dotfill}\\
}

\DeclareNewFootnote{A}
\DeclareNewFootnote{B}

\let\footnote\footnoteA

\begin{document}


\begin{center}
  {\huge Geometry-Calibrated DRO: Combating Over-Pessimism with Free Energy Implications} \\
  \vspace{.5cm}
  {\Large
    Jiashuo Liu$^{1,}$, Jiayun Wu$^{1}$, Tianyu Wang$^{2}$, Hao Zou$^3$, Bo Li$^4$, Peng Cui$^1$} \\
  \vspace{.2cm}
   $^1$Department of Computer Science and Technology, Tsinghua University\\
   $^2$Department of Industrial Engineering and Operations Research, Columbia University\\
   $^3$Zhongguancun Lab\\
   $^4$School of Economics and Management, Tsinghua University\\
    \vspace{.2cm}
  \texttt{liujiashuo77@gmail.com, cuip@tsinghua.edu.cn}\\
\end{center}


\else

\documentclass[twoside]{article}

\usepackage{aistats2024}
\usepackage{natbib}
\usepackage{amsmath}
\usepackage{amssymb}
\usepackage{mathtools}
\usepackage{amsthm}
\usepackage{xcolor}   
\usepackage{mathrsfs}
\usepackage{algorithm}
\usepackage{algorithmic}
\usepackage{graphicx}
\usepackage{subfig}
\usepackage{bbding}
\usepackage{multirow}
\usepackage{tabulary}
\usepackage{colortbl}
\usepackage{color}
\usepackage[utf8]{inputenc} 
\usepackage[T1]{fontenc}    
\usepackage{hyperref}       
\usepackage{url}            
\usepackage{booktabs}       
\usepackage{amsfonts}       
\usepackage{nicefrac}       
\usepackage{microtype}      
\usepackage{xcolor}         
\usepackage{amsmath}
\usepackage{amsthm}
\usepackage{amssymb}
\usepackage{mathrsfs}
\usepackage{algorithm}
\usepackage{algorithmic}
\usepackage{graphicx}
\usepackage{bbding}
\usepackage{multirow}
\usepackage{mathtools,color}
\newtheorem{assumption}{Assumption}[section]
\newtheorem{definition}{Definition}[section]
\newtheorem{proposition}{Proposition}[section]
\newtheorem{theorem}{Theorem}[section]
\newtheorem{lemma}{Lemma}[section]
\newtheorem{problem}{Problem}
\newtheorem*{example}{Example}
\newtheorem*{insight}{Insight}
\newtheorem*{remark}{Remark}
\newcommand{\tabincell}[2]{\begin{tabular}{@{}#1@{}}#2\end{tabular}}

\definecolor{babypink}{rgb}{0.96, 0.76, 0.76}
\definecolor{bittersweet}{rgb}{1.0, 0.44, 0.37}
\definecolor{blizzardblue}{rgb}{0.67, 0.9, 0.93}
\definecolor{brickred}{rgb}{0.8, 0.25, 0.33}
\definecolor{bubbles}{rgb}{0.91, 1.0, 1.0}
\definecolor{classicrose}{rgb}{0.98, 0.8, 0.91}
\definecolor{languidlavender}{rgb}{0.84, 0.79, 0.87}
\definecolor{pastelred}{rgb}{1.0, 0.41, 0.38}
\definecolor{lightpink}{rgb}{1.0, 0.71, 0.76}
%
%




\begin{document}

%

%

\twocolumn[

\aistatstitle{Geometry-Calibrated DRO: Combating Over-Pessimism with Free Energy Implications}

\aistatsauthor{ Author 1 \And Author 2 \And  Author 3 }

\aistatsaddress{ Institution 1 \And  Institution 2 \And Institution 3 } ]

\fi

\begin{abstract}
  Machine learning algorithms minimizing average risk are susceptible to distributional shifts. 
Distributionally Robust Optimization (DRO) addresses this issue by optimizing the worst-case risk within an uncertainty set. 
However, DRO suffers from over-pessimism, leading to low-confidence predictions, poor parameter estimations as well as poor generalization. 
In this work, we conduct a theoretical analysis of a probable root cause of over-pessimism: excessive focus on noisy samples. 
To alleviate the impact of noise, we incorporate data geometry into calibration terms in DRO, resulting in our novel Geometry-Calibrated DRO (GCDRO) {\bf \textit{for regression}}. 
We establish the connection between our risk objective and the Helmholtz free energy in statistical physics, and this free-energy-based risk can extend to standard DRO methods. 
Leveraging gradient flow in Wasserstein space, we develop an approximate minimax optimization algorithm with a bounded error ratio and elucidate how our approach mitigates noisy sample effects. 
Comprehensive experiments confirm GCDRO's superiority over conventional DRO methods.\let\thefootnote\relax\footnotetext{Short version appears at 37th Conference on Neural Information Processing Systems (NeurIPS 2023), Workshop on Distribution Shifts (DistShift).}
\end{abstract}

\input{data/1intro}

\input{data/2preliminary}

\input{data/3method}

\input{data/4experiment}

\input{data/5conclusion}

\newpage
\bibliography{example_paper}

 \def\usearxivstyle{1}
\ifdefined\usearxivstyle
\bibliographystyle{abbrvnat}
\else
\bibliographystyle{apalike}
\fi

\newpage
\input{data/6appendix.tex}

\end{document}

%% file: data/1intro.tex
\section{Introduction}
Machine learning algorithms with empirical risk minimization (ERM) have been shown to perform poorly under distributional shifts, especially sub-population shifts where substantial data subsets are underrepresented in the average risk due to their small sample sizes.
As an alternative, Distributionally Robust Optimization (DRO) \citep{namkoong2017variance,blanchet2019quantifying, blanchet2019robust,duchi2021learning,zhai2021doro,liu2021distributionally, gao2022distributionally, gao2022wasserstein} aims to optimize against the worst-case risk distribution within a predefined uncertainty set. 
This uncertainty set is centered around the training distribution, and generalization performance can be guaranteed when the test distribution falls within this set.

However, DRO methods have been found to experience the over-pessimism problem in practice \citep{hu2018does,zhai2021doro}~(\textit{i.e.}, low-confidence predictions, poor parameter estimations, and generalization), recent studies have sought to address this issue.
From the \emph{uncertainty set perspective},  \citet{DBLP:conf/wsc/BlanchetKMZ19, liu2021distributionally, liudistributionally} proposed data-driven methods to learn distance metrics from data. 
However, these approaches remain vulnerable to noisy samples, as demonstrated in Table \ref{tab:selection-bias}.
Recently, \citet{slowik2022distributionally, agarwal2022minimax} observed that DRO may overly focus on sub-populations with higher noise levels, leading to suboptimal generalization. 
Consequently, from the \emph{risk objective perspective}, they suggest incorporating calibration terms to mitigate this issue. 
Nevertheless, applicable calibration terms either require expert knowledge or are computationally intensive, and few practical algorithms have been proposed.

To devise a practical calibration term for DRO, we first aim to identify the root causes of over-pessimism, which we attribute to the excessive focus on noisy samples that frequently exhibit higher prediction errors.
For typical DRO methods \citep{namkoong2017variance, staib2019distributionally,duchi2021learning,liudistributionally}, based on a simple yet insightful linear example, we theoretically demonstrate that the variance of estimated parameters becomes substantially large when noisy samples have higher densities, in line with the empirical findings reported in \citep{zhai2021doro}.
Furthermore, we demonstrate that existing outlier-robust regression methods are not directly applicable for mitigating noisy samples in DRO scenarios where both noisy samples and distribution shifts coexist, highlighting the non-trivial nature of this problem.

In this work, inspired by the ideas in \citep{slowik2022distributionally, agarwal2022minimax}, we design calibration terms, $i.e.$, total variation and entropy regularization, to prevent DRO from excessively focusing on random noisy samples. 
In conjunction with the Geometric Wasserstein uncertainty set \citep{liudistributionally} utilized in our methods, these calibration terms effectively incorporate information from the data manifold, leading to improved regulation of the worst-case distribution in DRO. 
Specifically, during the optimization, the total variation term penalizes the variation of weighted prediction errors along the data manifold, preventing random noisy samples from gaining excessive densities. 
The entropy regularization term, also used in \citep{liudistributionally}, acts as a non-linear graph Laplacian operator that enforces the smoothness of the sample weights along the manifold. 
These calibration terms work together to render the worst-case distribution more \emph{reasonable} for DRO, leading to our Geometry-Calibrated DRO (GCDRO) approach.
We validate the effectiveness of our GCDRO on both simulation and real-world data.

Furthermore, from a statistical physics perspective, we demonstrate that our risk objective corresponds to the Helmholtz free energy, comprising three components: interaction energy, potential energy, and entropy. 
The free energy formulation generalizes typical DRO methods such as KL-DRO, $\chi^2$-DRO \citep{duchi2021learning}, MMD-DRO \citep{staib2019distributionally} and GDRO \citep{liudistributionally}.
This physical interpretation provides a novel perspective for understanding different DRO methods by drawing parallels between the worst-case distribution and the steady state in statistical physics, offering valuable insights.
From the free energy point of view, our GCDRO \emph{specifically addresses the interaction energy between samples to mitigate the effects of noisy samples}.
Motivated by the study of the Fokker-Planck equation (FPE, \citep{chow2017entropy, esposito2021nonlocal}), through gradient flow in the Geometric Wasserstein space, we derive an approximate minimax algorithm with a bounded error ratio $e^{-CT_{in}}$ after $T_{in}$ inner-loop iterations.
Our optimization method supports any quadratic form of interaction energy, potentially paving the way for designing more effective calibration terms for DRO in the future.

%% file: data/2preliminary.tex
\section{Preliminaries: Noisy Samples Bring Over-Pessimism in DRO}
\label{sec:preliminaries}
\textbf{Notations.}\quad 
$X\in\mathcal{X}$ denotes the covariates, $Y\in\mathcal{Y}$ denotes the target, $f_\theta(\cdot):\mathcal{X}\rightarrow\mathcal{Y}$ is the predictor parameterized by $\theta\in\Theta$. 
$\hat{P}_N$ denotes the empirical counterpart of distribution $P(X,Y)$ with $N$ samples, and $\bold{p}=(p_1,\dots, p_N)^T\in\mathbb{R}^N_+$ is the probability vector.
$[N]=\{1,2,\dots, N\}$ denotes the set of integers from 1 to $N$.
The random variable of data points is denoted by $Z=(X,Y)\in\mathcal{Z}$.
The random vector of $n$ dimension is denoted by $\vec{h}_n=(h_1,\dots,h_n)^T$.
$G_N=(V,E,W)$ denotes a finite weighted graph with $N$ nodes, where $V=[N]$ is the vertex set, $E$ is the edge set and $W=\{w_{ij}\}_{(i,j)\in E}$ is the weight matrix of the graph.
And $(x)_+=\max(x,0)$.

Distributionally Robust Optimization (DRO) is formulated as:
\begin{small}
\begin{equation}
\label{equ:noisemodel}
	\theta^*(P) = \arg\min\limits_{\theta\in\Theta}\sup\limits_{Q\in \mathcal{P}(P)}\mathbb{E}_Q[\ell(f_\theta(X),Y)]
\end{equation}	
\end{small}
where $\ell$ is the loss function (typically mean square error) and $\mathcal{P}(P)=\{Q: \text{Dist}(Q,P)\leq \rho\}$ denotes the $\rho$-radius uncertainty ball around the distribution $P$.
Different distance metrics derive different DRO methods, e.g., $f$-divergence DRO ($f$-DRO, \citet{namkoong2017variance, duchi2021learning}) with the Cressie-Read family of Rényi divergence, Wasserstein DRO (WDRO, \citet{sinha2017certifying,blanchet2019quantifying, blanchet2019robust,DBLP:conf/wsc/BlanchetKMZ19}), MMD-DRO \citep{staib2019distributionally} with maximum mean discrepancy, and Geometric DRO (GDRO, \citet{liudistributionally}) with Geometric Wasserstein distance.
Although DRO methods are designed to resist sub-population shifts, they have been observed to have poor generalization performances \citep{hu2018does, frogner2019incorporating, slowik2022distributionally} in practice, which is referred to as over-pessimism.

\begin{figure*}[t]
\vspace{-0.1in}
\centering
\begin{minipage}[t]{0.25\textwidth}
\centering
\includegraphics[width=\linewidth]{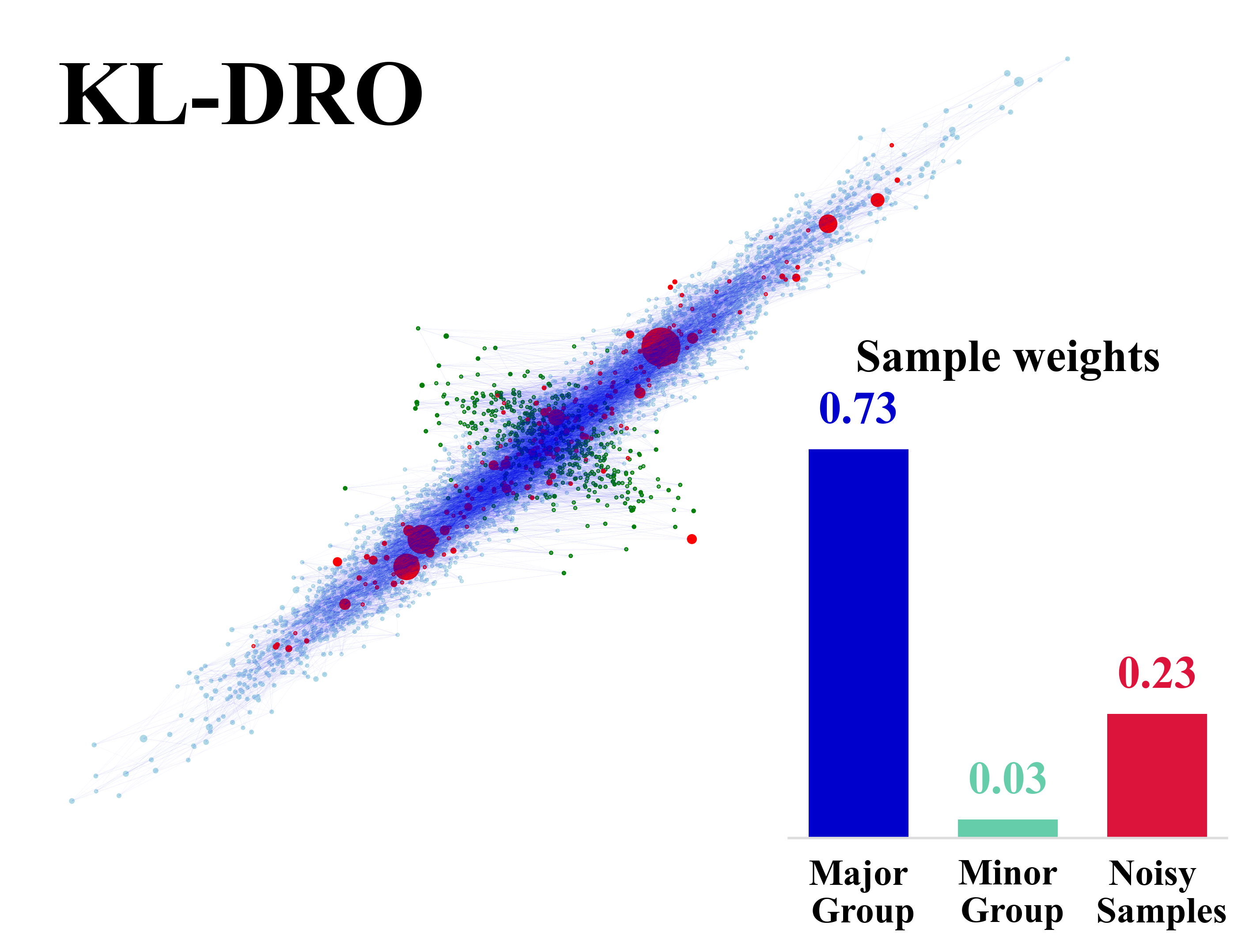}
\end{minipage}
\hskip -0.05in
\begin{minipage}[t]{0.25\textwidth}
\centering
\includegraphics[width=\linewidth]{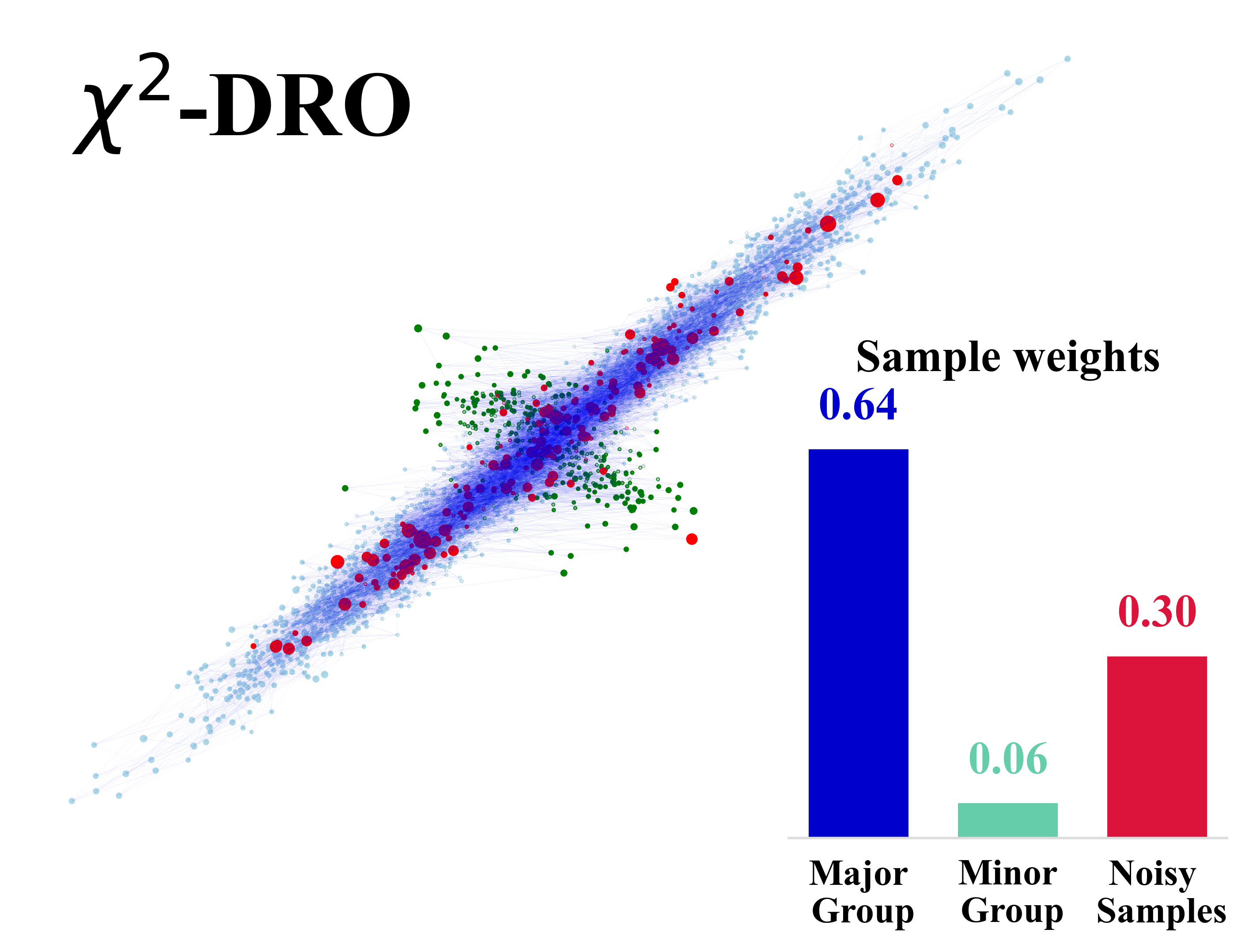}
\end{minipage}
\hskip -0.05in
\begin{minipage}[t]{0.25\textwidth}
\centering
\includegraphics[width=\linewidth]{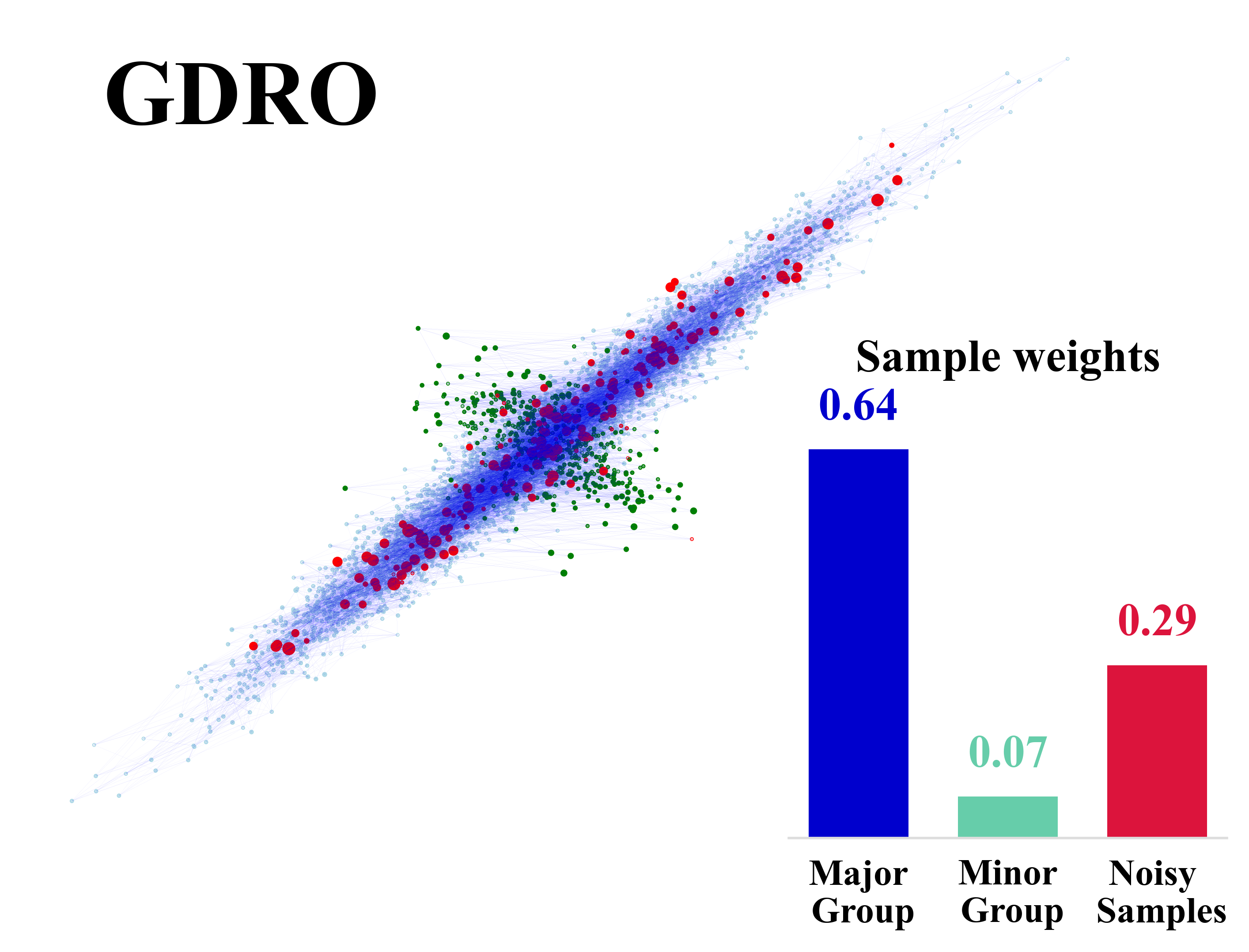}
\end{minipage}
\hskip -0.05in
\begin{minipage}[t]{0.25\textwidth}
\centering
\includegraphics[width=\linewidth]{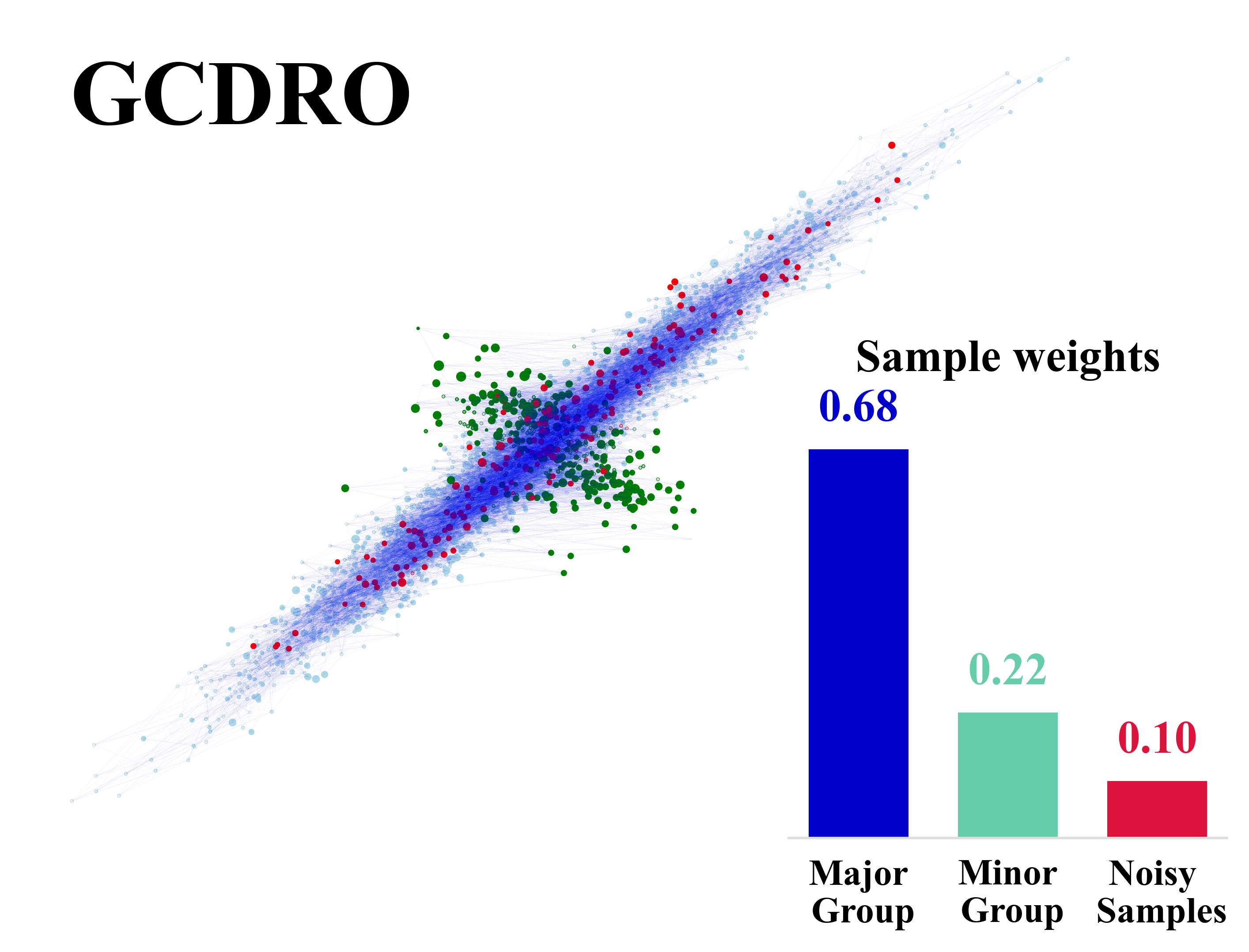}
\end{minipage}
\caption{Visualizing the Worst-Case Distribution for Different DRO Methods: We show the data manifold and sample weights for each point, where blue points represent the major group, green ones represent the minor group, and red ones are noisy samples. The bars display the total sample weights of different groups, and the \emph{original} group ratio is major (\textcolor{blue}{\bf 93.1\%}), minor (\textcolor{green}{\bf 4.9\%}), (noisy \textcolor{red}{\bf 2\%}).}
\label{fig: preliminary}
\vskip -0.15in
\end{figure*}

In this section, we identify one of the root causes of the over-pessimism of DRO: the \emph{excessive focus on noisy samples with typically high prediction errors}.\\ 
$\bullet$ We showcase DRO methods' excessive focus on noisy samples in practice and reveal their probability densities are linked to high prediction errors in worst-case distributions.\\
$\bullet$ Through a simple yet insightful regression example, we prove that such a phenomenon leads to high estimation variances and subsequently poor generalization performance.\\
$\bullet$ We demonstrate that existing outlier-robust regression methods are not directly applicable for mitigating noisy samples in DRO scenarios, emphasizing the non-trivial nature of this problem.

\textbf{Problem Setting}\quad Given the \emph{underlying} clean distribution $P_{clean}=(1-\alpha)P_{major}+\alpha P_{minor}, 0<\alpha<\frac{1}{2}$, the \textbf{goal of DRO can be viewed as achieving good performance across all possible sub-populations $P_{minor}$}. Denote the observed contaminated training distribution by $P_{train}$.
Based on Huber's $\epsilon$-contamination model \citep{huber1992robust}, we formulate $P_{train}$ as:
\begin{small}
\begin{equation}
\begin{aligned}
		P_{train} = (1-\epsilon)P_{clean} + \epsilon \tilde{Q} =
		\underbrace{(1-\epsilon)(1-\alpha)P_{major}}_{\text{major sub-population}}+
		\underbrace{(1-\epsilon)\alpha P_{minor}}_{\text{minor sub-population}} + \underbrace{\epsilon \tilde{Q}}_{\text{noisy sub-population}},
\end{aligned}
\end{equation}	
\end{small}
where $\tilde{Q}$ is an arbitrary \emph{noisy} distribution (typically with larger noise scale), $0<\epsilon<\frac{1}{2}$ is the noise level.
Note that the \emph{minor sub-population could represent any distribution with a proportion of $\alpha$ in $P$}. 
However, we explicitly specify it here to emphasize the distinction between our setting and the traditional Huber's $\epsilon$-contaminated setting, as the latter does \emph{not} take sub-population shifts into account.

\textbf{Empirical Observations.}\quad Following a typical regression setting \citep{duchi2021learning, liudistributionally}, we demonstrate the worst-case distribution of KL-DRO, $\chi^2$-DRO, and GDRO in Figure \ref{fig: preliminary}, where the size of each point is proportional to its density.
In this scenario, the underlying distribution $P$ comprises a known major sub-population (95\%, blue points) and a minor sub-population (5\%, green points). And the noise level $\epsilon$ in $P_{train}$ is $2\%$.
DRO methods are expected to upweight samples from minor sub-population to learn a model with uniform performances w.r.t. sub-populations.
However, from Figure \ref{fig: preliminary}, we could observe that KL-DRO, $\chi^2$-DRO and GDRO excessively focus on noisy samples, resulting in a noise level 10 to 15 times larger than the original.
This observation helps to explain their poor performance on this task (detailed results can be found in Table \ref{tab:selection-bias}).

\textbf{Theoretical Analysis.}\quad To support our observations, we first analyze the worst distribution of KL-DRO, $\chi^2$-DRO and GDRO, shedding light on the underlying reasons for this phenomenon.

\begin{proposition}[Worst-case Distribution]
\label{proposition: worst-case distribution}
	Let $\hat{Q}^*_N=(q^*_1, q^*_2, \dots, q^*_N)^T\in\mathbb{R}^N_+$ denotes the worst-case distribution,
	and $\ell(f_\theta(x_i),y_i)$ (\emph{abbr.} $\ell_i$) denotes the prediction error of sample $i\in[N]$.
	For different choices of $\text{Dist}(\cdot,\cdot)$ in $\mathcal{P}(P)=\{Q:\text{Dist}(Q,P)\leq \rho\}$, we have:\\
	$\bullet$ KL-DRO: $q^*_i/q^*_j \propto \exp(\ell_i-\ell_j)$; \\
    $\bullet$ GDRO's final state (gradient flow step $T\rightarrow \infty$): $q^*_i/q^*_j \propto \exp(\ell_i-\ell_j)$;\\	
    $\bullet$ $\chi^2$-DRO: $q^*_i/q^*_j = (\ell_i-\lambda)_+/(\ell_j-\lambda)_+$, and $\lambda\geq 0$ is the dual parameter independent of $i$.
\end{proposition}

Proposition \ref{proposition: worst-case distribution} demonstrates that for KL-DRO, $\chi^2$-DRO, and GDRO (large gradient flow step), the \emph{relative density} between samples is solely determined by their prediction errors, indicating that a larger prediction error results in a higher density. 
However, in our problem setting, samples from \emph{both} minor sub-population $P_{minor}$ \emph{and} noisy sub-population $\tilde{Q}$ exhibit high prediction errors.
The primary goal of DRO is to focus on the minor sub-population $P_{minor}$, but the presence of noisy samples in $\tilde{Q}$ significantly interferes with this objective and hurts model learning.
As shown in Figure \ref{fig: preliminary}, for KL-DRO, $\chi^2$-DRO and GDRO, noisy samples attract much density.
Intuitively, it is not surprising that an excessive focus on noisy samples can have a detrimental impact. 
As KL-DRO, $\chi^2$-DRO, and GDRO can be viewed as optimization within a weighted empirical distribution, we use the following simple example with the weighted least square model to demonstrate how this excessive focus on noisy samples can lead to high estimation variance, ultimately causing over-pessimism.

\begin{example}[Weighted Least Square]
\label{proposition:over-pessimism}
Consider the data generation process as $Y=kX+\xi$, where $X,Y\in\mathbb R$ and random noise $\xi$ satisfies $\xi \perp X$, $\mathbb E[\xi]=0$ and $\mathbb{E}[\xi^2]$ (\emph{abbr.} $\sigma^2$) is finite.
Assume that the training dataset $X_D$ consists of clean samples $\{ x_c^{(i)},y_c^{(i)} \}_{i \in [N_c]}$ and noisy samples $\{ x_o^{(i)}, y_o^{(i)}  \}_{i \in [N_o]}$ with $\sigma_c^2 < \sigma_o^2$. Consider the weighted least-square model $f(X) = \theta X$. Denote the sample weight of a clean sample $(x_c^{(i)}, y_c^{(i)})$ as $w_c^{(i)} \in \mathbb R_+, i \in [N_c]$, and the sample weight of a noisy sample $(x_o^{(i)}, y_o^{(i)})$ as $w_o^{(i)} \in \mathbb R_+, i \in [N_o]$ with $\sum_{i\in[N_c]}w_c^{(i)}+\sum_{i\in[N_o]}w_o^{(i)}=1$.
The variance of the estimator $\hat\theta$ is given by:
\begin{small}
\begin{align}
    \text{Var}[\hat\theta | X_D] =  \frac{\sum_{i=1}^{N_c} (w_c^{(i)})^2 (x_c^{(i)})^2 \sigma_c^2 + \sum_{i=1}^{N_o} (w_{o}^{(i)})^2 (x_o^{(i)})^2 \sigma_o^2}{\left[ {\sum_{i=1}^{N_c} w_c^{(i)} (x_c^{(i)})^2  + \sum_{i=1}^{N_o} w_o^{(i)} (x_o^{(i)})^2} \right]^2},
\end{align}
\end{small}
where $X_D = \{ x_c^{(i)} \}_1^{N_c}\cup \{ x_o^{(i)} \}_1^{N_o}$ are the sampled covariates in the dataset. 
Besides, the minimum variance is achieved if and only if  $\forall 1\leq i \leq N_c , 1\leq j\leq N_o,  w_o^{(j)}/w_c^{(i)} = \sigma_c^2 / \sigma_o^2 < 1$.
\end{example}
\vspace{-0.1in}
From the results, we make the following remarks:\\
$\quad\bullet$ If noisy samples have higher weights than clean samples (e.g., $w_o/w_c > 1$), the variance of the estimated parameter $\hat{\theta}$ will be larger, suggesting that the learned $\hat{\theta}$ could be significantly unstable.\\
$\quad \bullet$ In conjunction with Proposition \ref{proposition: worst-case distribution}, DRO methods tend to assign high weights to noisy samples, which can lead to unstable parameter estimation. While this example is relatively simple, this phenomenon aligns with the empirical findings in \citet{zhai2021doro}, which demonstrate that DRO methods can be quite unstable when confronted with label noise. 

\textbf{Relationship with Conventional Outlier-robust Regression.}\quad We would like to explain why conventional outlier-robust regression methods cannot be directly applied to our problem. The main challenge stems from the \emph{coexistence} of noisy samples and minor sub-populations, both of which typically exhibit high prediction errors, leading to a misleading worst-case distribution in DRO. 
Conventional outlier-robust regression methods \citep{diakonikolas2018algorithmic, klivans2018efficient, diakonikolas2022streaming} primarily focus on mitigating the effects of outliers without considering sub-population shifts. 
For instance, the $L_2$-estimation-error of outlier-robust linear regression is $\mathcal O(\epsilon\log(1/\epsilon))$ \citep{diakonikolas2018algorithmic}, where $\epsilon$ represents the noise level in Equation \ref{equ:noisemodel}. However, as analyzed in Proposition \ref{proposition: worst-case distribution} and demonstrated in Figure \ref{fig: preliminary}, during the optimization of DRO, the noise level $\epsilon$ significantly increases, rendering even outlier-robust estimation quite inaccurate.
Moreover, \cite{klivans2018efficient} propose finding a pseudo distribution with minimal prediction errors to avoid outliers (see Algorithm 5.2 in \citep{klivans2018efficient}). Nevertheless, this approach might inadvertently exclude minor sub-populations, which should be the focus under sub-population shifts, due to the main challenge: the \emph{coexistence} of noisy samples and minor sub-populations. \citet{zhai2021doro} incorporate this idea into DRO. Still, their method requires an implicit assumption that the prediction errors of noisy samples are higher than those of minor sub-populations, which does not always hold in practice.
And \citet{bennouna2022holistic} build the uncertainty set via two measures, KL-divergence and Wasserstein distance, leading to a combined approach of KL-DRO and ridge regression. 
Despite this, as we discussed earlier, DRO tends to increase the noise level in data, making it difficult to fix using ridge regression.

Based on the analysis above, we stress the importance of integrating more data-derived information. 
In pursuit of this, we propose to leverage the unique geometric properties that distinguish noisy samples from minor sub-populations to address this issue.

%% file: data/3method.tex
\section{Proposed Method}
\label{sec:method}
In this work, with a focus on regression, we introduce our Geometry-Calibrated DRO (GCDRO). 
The fundamental idea is to utilize data geometry to distinguish between random noisy samples and minor sub-populations. 
It is motivated by the fact that prediction errors for minor sub-populations typically exhibit local smoothness along the data manifold, a property that is not shared by noisy samples.

\textbf{Discrete Geometric Wasserstein Distance.}\quad We briefly revisit the definition of the discrete geometric Wasserstein distance.
Given a weighted finite graph $G_N=(V,E,W)$, the probability set $\mathscr{P}(G_N)$ supported on the vertex set $V$ is defined as $\mathscr{P}(G_N)=\{\bold{p}\in\mathbb R^N|\sum_{i=1}^N p_i=1, p_i\geq 0, \text{for }i \in V\}$, and its interior is denoted as $\mathscr{P}_o(G_N)$.
A velocity field $\bold{v}=(v_{ij})_{i,j\in V}\in \mathbb{R}^{N\times N}$ on $G_N$ is defined on the edge set $E$ satisfying that $v_{ij}=-v_{ji}$ if $(i,j)\in E$.
$\xi_{ij}(\bold{p})$ is a function interpolated with the associated nodes’ densities $p_i,p_j$.
The flux function $\bold{pv}\in\mathbb R^{N\times N}$ on $G_N$ is defined as $\bold{pv}:=(v_{ij}\xi_{ij}(\bold{p}))_{(i,j)\in E}$ and its divergence is defined as $\text{div}_{G_N}(\bold{pv}):= -(\sum_{j\in V: (i,j)\in E}\sqrt{w_{ij}}v_{ij}\xi_{ij}(\bold{p}))_{i=1}^N\in\mathbb{R}^N$.
Then for distributions $\bold{p}_0, \bold{p}_1 \in \mathscr{P}_o(G_N)$, the discrete geometric Wasserstein distance \citep{chow2017entropy, liudistributionally} is defined as:
\begin{small}
	\begin{equation}
	\begin{aligned}
	\label{equ:gw}
		\mathcal{GW}_{G_N}^2(\bold{p}_0,\bold{p}_1):=\inf\limits_v \bigg\{\int_0^1 \frac{1}{2}\sum_{(i,j)\in E}\xi_{ij}(\bold{p}(t))v_{ij}^2 dt\\
		\text{\quad s.t.}\frac{d\bold{p}}{dt}+\text{div}_{G_N}(\bold{pv})=0, \bold{p}(0)=\bold{p}_0, \bold{p}(1)=\bold{p}_1\bigg\}.
	\end{aligned}
	\end{equation}	
\end{small}
Equation \ref{equ:gw} computes the shortest (geodesic) length among all potential plans, integrating the total kinetic energy of the velocity field throughout the transportation process. 
A key distinction from the Wasserstein distance is that it only permits density to appear at the graph nodes.

\textbf{Formulation}\quad Given training dataset $D_{tr}=\{(x_i,y_i)\}_{i=1}^N$ and a finite weighted graph $G_N=(V,E,W)$ representing the inherent structure of sample covariates.
Denote the empirical marginal distribution as $\hat{P}_{X}$, the formulation of GCDRO is:
\vskip -0.1in
\begin{small}
\begin{equation}
\begin{aligned}
\label{equ:obj}
	\min_{\theta\in\Theta}\underbrace{\sup\limits_{\bold{q}:\mathcal{GW}_{G_N}^2(\hat{P}_{X},\bold{q})\leq \rho}}_{\text{Geometric Wasserstein set}} \bigg\{ \mathcal{R}_N(\theta,\bold{q}):=\sum_{i=1}^N q_i\ell(f_\theta(x_i),y_i)
	 - \underbrace{\frac{\alpha}{2}\cdot \sum_{(i,j)\in E}w_{ij}q_iq_j(\ell_i-\ell_j)^2}_{\text{Calibration Term \uppercase\expandafter{\romannumeral1}}}-\underbrace{\beta\cdot\sum_{i=1}^Nq_i\log q_i }_{\text{Calibration Term \uppercase\expandafter{\romannumeral2}}} \bigg\},
\end{aligned}
\end{equation}	
\end{small}
where $\rho$ is the pre-defined radius of the uncertainty set, $\ell_i$ is the loss on the $i$-th sample and $w_{ij}\in W$ denotes the edge weight between sample $i$ and $j$.
$\alpha$ and $\beta$ are hyper-parameters.

\textbf{Illustrations}.\quad In our formulation, for any distribution $\bold{q}$ within the uncertainty set, \\
\textbf{Calibration term \uppercase\expandafter{\romannumeral1}} ($\sum_{(i,j)\in E}w_{ij}q_iq_j(\ell_i-\ell_j)^2$) calculates the \emph{graph total variation} of prediction errors along the data manifold that is characterized by $G_N$.
Intuitively, when \emph{selecting the worst-case distribution}, this term imposes a penalty on distributions that allocate high densities to random noisy samples, as this allocation significantly amplifies the overall variation in prediction errors.
Conversely, this term does not penalize distributions that allocate high densities to minor sub-populations, as their errors are smooth and have a relatively small impact on the total variation along the manifold.
This differing phenomenon arises from the distinct geometric properties of random noisy samples and minor sub-populations, as samples from the latter typically cluster together on the data manifold.
Further, \emph{during the optimization of model parameter $\theta$}, this term acts like a variance term, resulting in a quantile-like risk objective, which helps to mitigate the effects of outliers.\\
\textbf{Calibration term \uppercase\expandafter{\romannumeral2}} ($\sum_{i=1}^Nq_i\log q_i$) represents the negative entropy of distribution $\bold{q}$. As discussed in Section \ref{subsec:optimization}, during optimization, this term transforms into a non-linear \emph{graph Laplacian operator} that encourages sample weights to be smooth along the manifold, avoiding extreme sample weights in the worst-case distribution.

\begin{table*}[t]
\vskip -0.1in
\caption{Free energy implications of some DRO methods. $\Delta_N$ denotes the $N$-dimensional simplex, $\eta$ in marginal DRO is the dual parameter.}
\label{tab:unify}
\centering\resizebox{0.8\textwidth}{2.3cm}{
\begin{tabular}{lccccccc}
\toprule
\multicolumn{1}{c}{\multirow{2}{*}{Method}} & \multicolumn{3}{c}{Energy Type}   & \multicolumn{4}{c}{Specific Formulation} \\ \cmidrule(lr){2-4}\cmidrule(lr){5-8}
\multicolumn{1}{c}{}                        & Interaction & Potential & Entropy & $K$   & $V$   & $H[\bold{q}]$   & $\mathscr P$  \\ \midrule
KL-DRO                                      &\XSolidBrush             &  \CheckmarkBold &\CheckmarkBold         & -       &$-\vec{\ell}$       &$H[\bold{q}]$          &  $\Delta_N$              \\
$\chi^2$-DRO                                &\CheckmarkBold             &\CheckmarkBold           &\XSolidBrush         &       $\lambda I$ &$-\vec{\ell}$       & -         &   $\Delta_N$            \\
MMD-DRO                                     &\CheckmarkBold             &\CheckmarkBold           &\XSolidBrush         &       \begin{tabular}[c]{@{}c@{}}Kernel Gram\\ Matrix $K$\end{tabular}& $-\vec{\ell}-\frac{2\lambda}{N}K^\top\bold{1}$      & -         & $\Delta_N$              \\
Marginal $\chi^2$-DRO                                &\XSolidBrush             &\CheckmarkBold           &\XSolidBrush         & -      & $-(\vec{\ell}-\eta)_+$      &    -      & \begin{tabular}[c]{@{}c@{}}$\Delta_N$ with Hölder\\ continuity\end{tabular}              \\
GDRO                                        &\XSolidBrush             &\CheckmarkBold           &\CheckmarkBold         & -       & $-\vec{\ell}$       & $H[\bold{q}]$          & \begin{tabular}[c]{@{}c@{}}Geometric\\ Wasserstein Set \end{tabular}               \\
GCDRO                                       &\CheckmarkBold             &\CheckmarkBold           &\CheckmarkBold         & \begin{tabular}[c]{@{}c@{}}Interaction\\ Matrix $K$\end{tabular}      & $-\vec{\ell}$      &  $H[\bold{q}]$         & \begin{tabular}[c]{@{}c@{}}Geometric\\ Wasserstein Set \end{tabular}   \\ \bottomrule            
\end{tabular}}
\vskip -0.1in
\end{table*}
\subsection{Free Energy Implications on Worst-case Distribution}
\label{subsec:physical}
We first demonstrate the free energy implications of our risk objective $\mathcal R_N(\theta,\bold{q})$.
Intuitively, the change of sample weights across $N$ samples (the inner maximization problem of $\mathcal R_N(\theta, \bold{q})$) can be analogously related to the dynamics of particles in a system, wherein the concentration of densities coincides with the aggregation of particle masses at $N$ distinct locations (in the case of infinite samples, these locations converge to the data manifold). 
As a result, a deeper understanding of the steady state in a particle system can offer \emph{valuable insights into the worst-case distribution} for DRO.

Building on this analogy, we can dive deeper into the physics of particle interactions. 
When particles exist within a potential energy field, they are subject to external forces. 
Simultaneously, there are interactions among the particles themselves, leading to a constant state of motion within the system. 
In statistical physics, a key point of interest is identifying when a system reaches a steady state. 
In a standard process like the reversible isothermal process, it is established that spontaneous reactions consistently move in the direction of decreasing \emph{Helmholtz free energy} \citep{fu1990physical, reichl1999modern, friston2010free}, which consists of interaction energy, potential energy and the negative entropy:
\begin{small}
\begin{equation}
\label{equ:free-energy}
\begin{aligned}
	\mathcal E(\bold{q})  = \underbrace{\bold{q}^\top K\bold{q}}_{\text{Interaction Energy}} + \underbrace{\bold{q}^\top V}_{\text{Potential Energy}}-\underbrace{\beta\sum_{i=1}^N (-q_i\log q_i)}_{\text{Temperature}\times \text{Entropy}}
	= -\mathcal R_N(\theta,\bold{q}).
\end{aligned}
\end{equation}	
\end{small}
By taking $V=-\vec{\ell}$ and $K_{ij}=\frac{\alpha}{2} w_{ij}(\ell_i-\ell_j)^2$ for $(i,j)\in E$, our risk objective is a special case of Helmholtz free energy, where the potential energy of sample $i$ is $-\ell_i q_i$ and the interaction energy between sample $i$ and $j$ is $\frac{\alpha}{2}w_{ij}(\ell_i-\ell_j)^2q_iq_j$.
Specifically, such mutual interactions can manifest as \emph{repulsive forces between adjacent particles}, thereby preventing the concentration of mass in locations where local prediction errors are significantly high.
And this explains from a physical perspective why our calibration term \textbf{\uppercase\expandafter{\romannumeral1}} could mitigate random noisy samples.

Additionally, Proposition \ref{proposition:unify} offers physical interpretations to comprehend the worst-case distribution of various DRO methods. 
We make some remarks: (1) current DRO methodologies, except MMD-DRO, do not explicitly formulate the interaction term between samples in their design considerations ($\chi^2$-DRO does not involve interaction between samples), despite the corresponding interaction energy between particles being a common phenomenon in physics; (2) MMD-DRO simply uses kernel gram matrix for interaction and lacks efficient optimization algorithms; (3) by \emph{considering this interaction energy}, our proposed GCDRO is capable of mitigating the impacts of random noisy samples.

\begin{proposition}[Free Energy Implications]
\label{proposition:unify}
	The dual reformulations of some typical DRO methods are equivalent to the free-energy-based minimax problem $\min_{\theta\in\Theta,\lambda\geq 0}\max_{\bold{q}\in\mathscr{P}}\bigg\{\lambda\rho-\mathcal E(\bold{q},\theta, \lambda)\bigg\}$
	with different choices of $\mathscr{P},\rho$ and $K,V,H[q]$ in the free energy $\mathcal E$.
	Details are shown in Table \ref{tab:unify}.
\end{proposition}
Through free energy, we could understand the type of energy or steady state that DRO methods strive to achieve, and design better interaction energy terms in DRO.
Moreover, our optimization, as outlined in Section \ref{subsec:optimization}, could accommodate multiple quadratic forms of interaction energy.

\subsection{Optimization}
\label{subsec:optimization}
Then we derive an approximate minimax optimization for our GCDRO.
For the \emph{inner maximization} problem, we approximately deal with it via the gradient flow of $-\mathcal{R}_N(\theta,Q)$ w.r.t. $Q$ in the geometric Wasserstein space $(\mathscr{P}_o(G_N), \mathcal{GW}_{G_N})$.
We show that the error rate is $\mathcal O(e^{-CT_{in}})$ after $T_{in}$ iterations inner loop, which gives a nice approximation.

We denote the \emph{Continuous gradient flow} as $\bold{q}:[0,T]\rightarrow \mathscr{P}_o(G_N)$, the probability density of sample $i$ at time $t$ is abbreviated as $q_i(t)$, and the \emph{Time-discretized gradient flow} with time step $\tau$ as $\hat{\bold{q}}_{\tau}$.
For inner maximization, we utilize the $\tau$-time-discretized gradient flow \citep{optimaltransport} for $-\mathcal{R}_N(\theta,\bold{q})$ in the geometric Wasserstein space $(\mathscr{P}_o(G_N), \mathcal{GW}_{G_N}^2)$ as:
\begin{small}
\begin{equation}
\label{equ:q}
	\hat{\bold{q}}_\tau(t+\tau) = \mathop{\text{argmax}}\limits_{\bold{q}\in\mathscr{P}_o(G_N)}\mathcal{R}_N(\theta,\bold{q})-\frac{1}{2\tau}\mathcal{GW}_{G_N}^2(\hat{\bold{q}}_\tau(t),\bold{q}).
\end{equation}	
\end{small}
The gradient of $\bold{q}$ in Equation \ref{equ:q} is given as (when $\tau\rightarrow 0$):
\begin{small}
\begin{equation}
\begin{aligned}
\label{equ:gradient}
	\frac{dq_i}{dt}=\sum_{(i,j)\in E} w_{ij}\xi_{ij}\bigg(\bold{q},\ \ \ell_i-\ell_j+\beta(\log q_j-\log q_i)+
	\alpha\big(\sum_{h\in N(j)}(\ell_h-\ell_j)^2w_{jh}q_h-\sum_{h\in N(i)}(\ell_h-\ell_i)^2w_{ih}q_h \big)\bigg)	,
\end{aligned}
\end{equation}	
\end{small}
where $E$ is the edge set of $G_N$, $w_{ij}$ is the edge weight between node $i$ and $j$, $N(i)$ denotes the set of neighbors of node $i$, $\ell_i$ denotes the loss of sample $i$, and $\xi_{ij}(\cdot,\cdot):\mathscr{P}(G_N)\times\mathbb R\rightarrow \mathbb R$ is: 
\begin{small}
\begin{equation}
	\xi_{ij}(\bold{q},v):= v\cdot\big(\mathbb I(v>0)q_j + \mathbb I(v\leq 0)q_i\big), v\in\mathbb R,
\end{equation}	
\end{small}
which is the \emph{upwind interpolation} commonly used in statistical physics and guarantees that the probability vector $\bold{q}$ keeps positive.
From the gradient, we could see that the entropy regularization acts as a non-linear graph Laplacian operator to make the sample weights smooth along the manifold.
In our algorithm, we fix the steps of the gradient flow to be $T_{in}$ and prove that the error ratio is $e^{-CT_{in}}$ compared with the \emph{ground-truth} worst-case risk $\mathcal R_N(\theta,\bold{q}^*)$ constrained in an $\rho(\theta,T_{in})$-radius ball.

\begin{proposition}[Approximation Error Ratio]
	\label{proposition:error-rate}
	Given the model parameter $\theta$, denote the distribution after time $T_{in}$ as $\bold{q}^{T_{in}}(\theta)$, and the distance to training distribution $\hat{P}_{X}$ as $\rho(\theta, T_{in}):=\mathcal{GW}_{G_N}^2(\hat{P}_{X}, \bold{q}^{T_{in}}(\theta))$ (\emph{abbr.} $\rho(\theta)$).
    Assume $\mathcal R_N(\theta,\bold{q})$ is convex w.r.t $\bold{q}$.
	Then define the ground-truth worst-case distribution $q^*(\theta)$ within the $\rho(\theta)$-radius ball as:
	\begin{small}
	\begin{equation}
	\label{equ:p-star}
		\bold{q}^*(\theta) := \arg\sup\limits_{\bold{q}:\mathcal{GW}_{G_N}^2(\hat{P}_{X},\bold{q})\leq \rho(\theta)} \mathcal R_N(\theta,\bold{q}).
	\end{equation}	
	\end{small}
	The upper bound of the error rate of the objective function $\mathcal{R}_N(\theta, \bold{q}^{T_{in}})$ satisfies:
	\begin{small}
	\begin{align}
	\label{equ:error-rate}
		\frac{\mathcal{R}_N(\theta, \bold{q}^*) -\mathcal{R}_N(\theta, \bold{q}^{T_{in}})}{\mathcal{R}_N(\theta, \bold{q}^*)-\mathcal{R}_N(\theta, \hat{P}_{X})} < e^{-CT_{in}},
		\quad C=2m\lambda_{\text{sec}}(\hat{L})\lambda_{\text{min}}(\nabla^2\mathcal R_N)\frac{1}{(r+1)^2} > 0,
	\end{align}	
	\end{small}
	where $\hat{L}$ is the Laplacian matrix of $G_N$.
	$\lambda_{\text{sec}}, \lambda_{\text{min}}$ are the second smallest and smallest eigenvalue, $m,r$ are constants depending on $\mathcal R_N, G_N, \beta$.
\end{proposition}
We make some remarks:\\
$\quad \bullet$ For the assumption that $\mathcal R_N$ is convex w.r.t. $\bold{q}$, the Hessian is given by $\nabla^2\mathcal R_N=\beta\text{diag}(1/q_1,...,1/q_N)+2K$. 
Since $K$ is a sparse matrix whose nonzero elements in each row is far smaller than $N$, it is easily satisfied in empirical settings that the Hessian matrix $\nabla^2\mathcal R$ is diagonally dominant and thus positive definite, making the inner maximization concave w.r.t $\bold{q}$.\\
$\quad\bullet$ During the optimization, our algorithm finds an approximate worst-case distribution that is close to the ground-truth one within a $\rho(\theta)$-radius uncertainty set. Our robustness guarantee is similar to \citet{sinha2017certifying} (see Equation 12 in \citet{sinha2017certifying}).\\
$\quad\bullet$ The error ratio is $e^{-CT_{in}}$, enabling to find a nice approximation efficiently with finite $T_{in}$ steps.

\subsection{Mitigate the Effects of Random Noisy Samples}
\label{subsec:theoretical}
Finally, we prove that our GCDRO method effectively de-emphasizes 'noisy samples' with locally non-smooth prediction errors.
Due to the challenge of assessing intermediate states in gradient flow, we focus on its final state (as $T_{in}\rightarrow \infty$). 

For the worst-case distribution $q^*$, we denote the density ratio between samples as $\gamma(i,j):=q^*_i/q^*_j$.
In sensitivity analysis, when \emph{only} sample $i$ is perturbed with label noises, we denote the density ratio in the new worst-case distribution $\tilde{q^*}$ as $\gamma^{\text{noisy}}(i,j):=\tilde{q^*_i}/\tilde{q^*_j}$.
The sample weight sensitivity $\xi(i,j)$ is defined as $\xi(i,j)=\log \gamma^{\text{noisy}}(i,j)-\log\gamma(i,j)$, which measures how much density ratio changes under perturbations on one sample.
Larger $\xi(i,j)$ indicates larger sensitivity to noisy samples.

\begin{proposition}
\label{proposition:label-noise}
Assume $\ell_i^{\text{noisy}}-\ell_i \geq 2(\frac{\sum_{k\in N(i)}q^*_kw_{ik}\ell_k}{\sum_{k\in N(i)}q^*_kw_{ik}}-\ell_i)$ which is locally non-smooth.
For any $\alpha>0$ (in Equation \ref{equ:obj}), we have $\xi_{\text{GCDRO}}<\xi_{\text{GDRO}}$. Furthermore, there exists $M>0$ such that for any $\alpha>M$, we have $\xi_{\text{GCDRO}}(i,j)<0<\min\{\xi_{\chi^2-\text{DRO}}(i,j), \xi_{\text{GDRO}}(i,j)(=\xi_{\text{KL-DRO}}(i,j))\}$, indicating that GCDRO is not sensitive to locally non-smooth noisy samples.
\end{proposition}
In practice, we do a grid search over $\alpha \in [0.1,10]$ on an independent held-out validation dataset to select the best $\alpha$.
The complexity of gradient flow scales \emph{linearly} with sample size.


%% file: data/4experiment.tex
\section{Experiments}
In this section, we test the empirical performances of our proposed GCDRO on simulation data and real-world \emph{regression} datasets with natural distributional shifts.
As for the baselines, we compare with empirical risk minimization (ERM), WDRO, two typical $f$-DRO methods, including KL-DRO, $\chi^2$-DRO \citep{duchi2021learning}, GDRO \citep{liudistributionally}, HRDRO \citep{bennouna2022holistic}  and DORO \citep{zhai2021doro}, where HRDRO and DORO are designed to mitigate label noises.

\begin{table*}[h]
\vskip -0.1in
\caption{Results on the simulation data. We report the average root mean square errors (RMSE) over 5 runs, excluding the small standard deviations.}
\label{tab:selection-bias}
\centering
\resizebox{\textwidth}{1.7cm}{
\begin{tabular}{@{}lcccccccccc@{}}
\toprule
             & \multicolumn{5}{c}{\large Weak Label Noise (noise level $0.5\%$)}                                                                                 & \multicolumn{5}{c}{\large Strong Label Noise (noise level $5\%$)}                                                                               \\ \cmidrule(lr){2-6}\cmidrule(lr){7-11}
             & Train (major) & Train (minor) & Test Mean & Test Std & \begin{tabular}[c]{@{}c@{}}Parameter\\ Est Error\end{tabular} & Train (major) & Train (minor) & Test Mean & Test Std & \begin{tabular}[c]{@{}c@{}}Parameter\\ Est Error\end{tabular} \\
ERM          & 0.337         & 0.850         & 0.598     & 0.264    & 0.423                                                         & 0.368         & 0.855         & 0.599     & 0.243    & 0.431                                                         \\
WDRO         & 0.337         & 0.851         & 0.589     & 0.292    & 0.424                                                         & 0.368         & 0.857         & 0.600     & 0.268    & 0.432                                                         \\
$\chi^2$-DRO & 0.596         & 0.765         & 0.680     & 0.088    & 0.447                                                         & 1.072         & 0.708         & 0.875     & 0.193    & 0.443                                                         \\
KL-DRO       & 0.379         & 1.616         & 0.974     & 0.660    & 0.886                                                         & 0.468         & 1.683         & 1.037     & 0.621    & 0.913                                                         \\
HRDRO       & 0.325         & 1.298     & 0.794         & 0.516    & 0.693 & 0.330 & 1.343 & 0.801 & 0.522 & 0.694\\
DORO         & 0.347         & 0.793         & 0.565     & 0.230    & 0.384                                                         & 0.334         & 0.919         & 0.611     & 0.295    & 0.449                                                         \\
GDRO         & 0.692         & 0.516         & 0.605     & 0.094    & 0.198                                                         & 0.618         & 0.752         & 0.677     & 0.063    & 0.421                                                         \\
GCDRO        & 0.411         & 0.554         & \bf 0.482     & \bf 0.070    & \bf 0.190                                                         & 0.494         & 0.591         & \bf 0.540     & \bf 0.044    & \bf 0.268                                                         \\ \bottomrule
\end{tabular}}
\end{table*}

\subsection{Simulation Data}

\textbf{Data Generation.}\quad We design simulation settings with both sub-population shifts and noisy samples.
The input covariates $X=[S,U,V]^T\in\mathbb R^{10}$ consist of stable covariates $S\in\mathbb R^5$, irrelevant ones $U\in\mathbb R^4$ and the unstable covariate $V\in\mathbb R$:
\begin{small}
\begin{align}
	&[S,U]\sim \mathcal N(0, 2\mathbb I_9), Y = \theta_S^TS+0.1S_1S_2S_3 + \mathcal{N}(0, 0.5),\\ 
	&V \sim \text{Laplace}(\text{sign}(r)\cdot Y, 1/5\ln |r|),
\end{align}	
\end{small}
where $\theta_S\in\mathbb R^5$ is the coefficients of the true model, $|r|>1$ is the adjustment factor for each sub-population, and $\text{Laplace}(\cdot,\cdot)$ denotes the Laplace distribution.
From the data generation, the relationship between $S$ and $Y$ stays invariant under different $r$, $U\perp Y$, while the relationship between $V$ and $Y$ is controlled by $r$, which \emph{varies across sub-populations}.
Intuitively, $\text{sign}(r)$ controls whether the spurious correlation $V$-$Y$ is positive or negative. 
And $|r|$ controls the strength of the spurious correlation: the larger $|r|$ is, the stronger the spurious correlation is.
Furthermore, in order to conform to real data which are naturally assembled with label noises \citep{zhai2021doro}, we introduce label noises by an $\epsilon$ proportion of labels as $Y' \sim \mathcal{N}(0, \text{Std}(Y))$. 
$\epsilon$ controls the noise level.

\textbf{Settings.}\quad In training, we generate 9,500 points with $r=1.9$ (\emph{majority}, strong positive spurious correlation $V$-$Y$) and 500 points with $r=-1.3$ (\emph{minority}, weak negative spurious correlation $V$-$Y$).
In testing, we vary $r\in\{3.0, 2.3, -1.9,-2.7\}$ to simulate different spurious correlations $V$-$Y$.
We use \emph{linear model} with mean square error (MSE) and report the prediction root-mean-square errors (RMSE) for each sub-population, the mean and standard deviation of prediction errors among all testing sub-populations.
Also, we report the parameter estimation errors $\|\hat{\theta}-\theta^*\|_2$ of all methods ($\theta^*=(\theta_S^T,0,\dots,0)^T$).
The results over 10 runs are shown in Table \ref{tab:selection-bias}.

\begin{figure*}[t]
  \subfloat[\texttt{Bike Dataset}]{\includegraphics[width=0.32\textwidth]{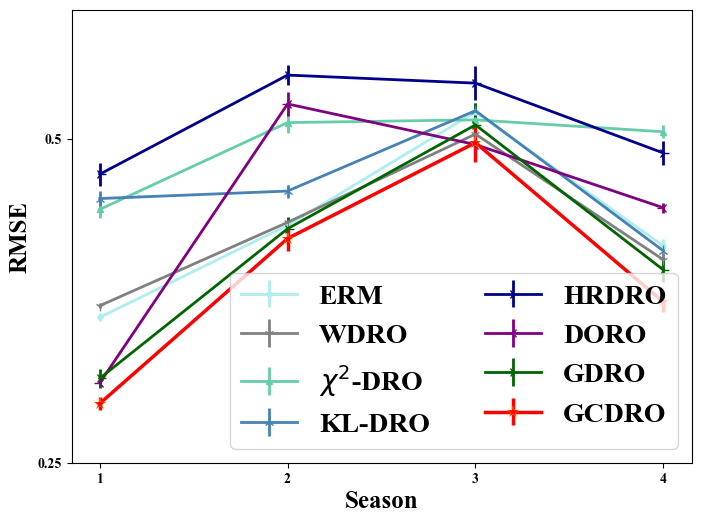}\label{fig:bike}}
  \hfill
  \subfloat[\texttt{House Dataset}]{\includegraphics[width=0.32\textwidth]{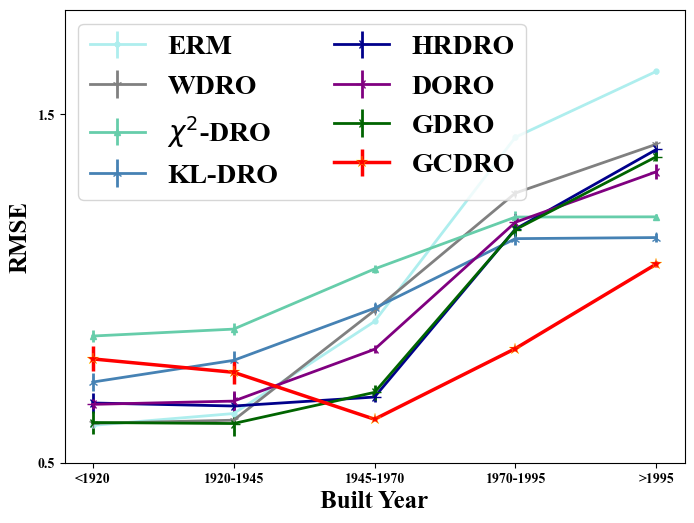}\label{fig:house}}
  \hfill
  \subfloat[\texttt{Temperature Dataset}]{\includegraphics[width=0.32\textwidth]{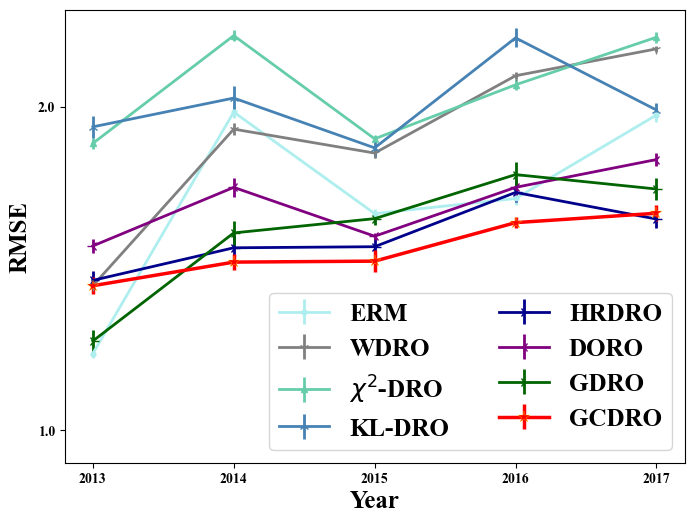}\label{fig:temperature}}
  
  \caption{Results (over 5 runs) of real-world datasets with natural shifts. We do not manually add label noises here, since real-world datasets intrinsically contain noises.}
 \vskip -0.2in
\end{figure*}


\textbf{Analysis.}\quad From Table \ref{tab:selection-bias}, (1) compared with ERM, all typical DRO methods, especially $\chi^2$-DRO and KL-DRO, are strongly affected by label noises. (2) Although DORO is designed to mitigate outliers, it does not perform well under strong noises ($\kappa=5\%$), because it relies on the assumption that noisy points have the largest prediction errors, which does not always hold. (3) Our proposed GCDRO outperforms all baselines under different strengths of label noises, which demonstrates its effectiveness. (4) Compared with GDRO, we could see that our \emph{calibration terms} in Equation \ref{equ:obj} is effective to mitigate label noises.
From Figure \ref{fig: preliminary}, the worst-case distribution of our GCDRO \emph{significantly upweighs on the minority} (green points) and does not put much density on the noisy data (red points), while the others put much higher weights on the noisy samples and perform poorly.

\subsection{Real-world Data}
We use three real-world regression datasets with natural distributional shifts, including bike-sharing prediction, house price, and temperature prediction. 
For all these experiments, we use a two-layer \emph{MLP} model with mean square error (MSE).
We use the Adam optimizer \cite{adam} with the default learning rate $1e-3$.
And all methods are trained for $5e3$ epochs.\\

\textbf{Datasets.}\quad (1) \textbf{Bike-sharing} dataset \citep{Dua:2019} contains the daily count of rental bikes in the Capital bike-sharing system with the corresponding 11 weather and seasonal covariates.
The task is to predict the count of rental bikes of \emph{casual users}.
Note that the count of casual users is likely to be more \emph{random and noisy}, which is suitable to verify the effectiveness of our method.
We split the dataset according to the season for natural shifts.
In the training data, the ratio of four seasons' data is $9:7:5:3$.
We test on the rest of the data and report the prediction error of each season.\\
(2) \textbf{House Price} dataset\footnote{https://www.kaggle.com/c/house-prices-advanced-regression- techniques/data} contains house sales prices from King County, USA.
The task is to predict the transaction price of the house via 17 predictive covariates such as the number of bedrooms, square footage of the house, etc.
We divide the data into 5 sub-populations according to the built year of each house with each sub-population covering a span of 25 years.
In training, we use data from the first group (built year $<1920$) and report the prediction error for each testing group.\\
(3) \textbf{Temperature} dataset \citep{Dua:2019} is largely composed of the LDAPS model's next day's forecast data, in-situ maximum and minimum temperatures of present-day, and geographic auxiliary variables in South Korea from 2013 to 2017. 
The task is to predict the next-day's maximum air temperatures based on the 22 covariates.
We divide the data into 5 groups corresponding with 5 years.
In the training data, the ratio of five years' data is $9:7:5:3:1$.
We test on the rest of the data and report the prediction error of each year.
More details could be found in Appendix.

\textbf{Analysis.}\quad
(1) From the results in Figure \ref{fig:bike}, we could see that the performances of ERM drop a lot under distributional shifts, and DRO methods have better performance as well as robustness.
(2) Our proposed GCDRO outperforms all baselines under strong shifts, with the most stable performances under natural distributional shifts.
(3) As for the $k$NN graph's fitting accuracy of the data manifold, we visualize the learned manifold in Appendix and we could see that the learned $k$NN graph fits the data manifold well.
Besides, we show in Appendix that the performances of our GCDRO are relatively stable across different choices of $k$.
Also, our GCDRO only needs the input graph $G_N$ to represent the data structure and \emph{any manifold learning or graph learning} methods could be plugged in to give a better estimation of $G_N$.

%% file: data/5conclusion.tex
\section{Future Directions}
Our work deals with the over-pessimism in DRO via geometric calibration terms and provides free energy implications.
The high-level idea could inspire future research on (1) relating free energy with DRO; (2) designing more reasonable calibration terms in DRO; (3) incorporating data geometry in general risk minimization algorithms.
We hope this work could help to make DRO methods more effective in practice.
And future improvements may be extend this method to classification scenarios with more complicated data like images and languages.

%% file: data/6appendix.tex
\appendix
\onecolumn
\section{Implementation}
For our GCDRO, $G_N$ is constructed as a $k$-nearest neighbor ($k$NN) graph from training data \emph{once and for all} \textbf{\textit{only at the initialization step}}.
For large-scale datasets, we use NN-Descent to estimate the $k$NN graph with an almost linear complexity of $\bf\mathcal O(N^{1.14})$.
Since the sample weights are transferred along the edges of the graph, the simulation of gradient flow can be implemented similarly to message propagation with DGL package \citep{DBLP:journals/corr/abs-1909-01315}, which \textbf{\textit{scales linearly with sample size}} and enjoys \textit{parallelization by GPU}.
The implementation above ensures the adaptability to large-scale data.

\section{Improvements of our work.}
\label{appendix:related}
In Section \ref{sec:preliminaries}, we have introduced the typical DRO methods in detail and demonstrated the over-pessimism problem.
Here we compare our work with several DRO works and clarify their differences.\\
(1) With MMD-DRO: MMD-DRO \citep{staib2019distributionally} also has a quadratic term in its dual reformulation, while \cite{staib2019distributionally} focuses on the equivalence between MMD-DRO and Hilbert norms and there is no efficient or applicable algorithm yet. 
Further, it remains the risk objective unchanged (the quadratic term is from MMD distance) and just uses the Gaussian RBF kernel.
Our work firstly incorporates the data geometry into the design of the calibration term and demonstrates its relationship with Helmholtz free energy, and we propose an applicable algorithm that could be used under deep models.\\
(2) With GDRO: GDRO \citep{liudistributionally} uses the discrete geometric Wasserstein distance to build the uncertainty set, and intuitively demonstrates its superiority.
Our work theoretically analyzes the over-pessimism problem and attributes the cause of over-pessimism to the excessive focus on noisy samples in DRO.
And for the risk objective function, our work further introduces the graph total variation term to mitigate the effects of noisy samples, which is theoretically justified and empirically verified.
From our results, GDRO is heavily affected by noisy samples, while our GCDRO has a much better performance.
Further, this work relates the newly-proposed risk objective to the Helmholtz free energy and unifies some typical DRO methods into it, which is a new perspective to view DRO methods and could inspire future research.\\
(3) With DORO: DORO \citep{zhai2021doro} proposes to dismiss data samples with the top losses and then performs DRO, and we compare with it in our experiments.
Theoretically, this method relies on the implicit assumption that noisy samples must have larger prediction errors than hard clean samples.
However, this assumption does not always hold, and as shown in our experiments, it has some effects but does not work very well.

\section{Why uses $k$NN graph?}

\begin{figure}[h]
	\centering\includegraphics[width=0.6\linewidth]{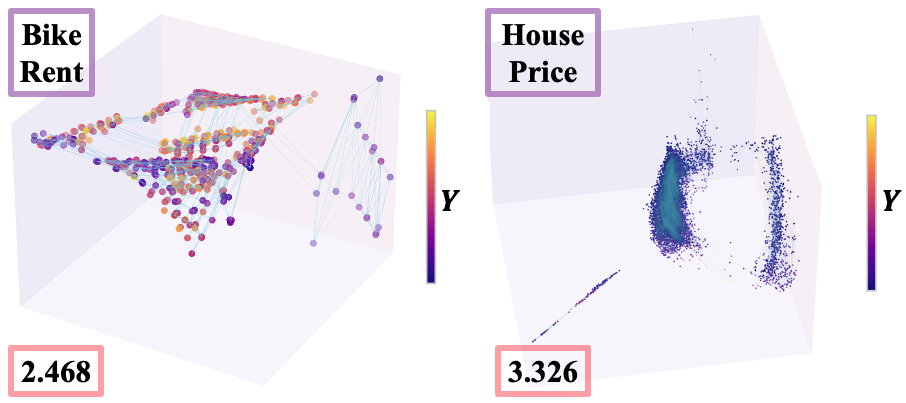}
	\caption{Visualization of the 3-dimensional manifold of the tabular datasets, and the numbers in the lower left represent the intrinsic dimension according to \cite{DBLP:conf/nips/LevinaB04}}
	\label{fig: manifold}
	\vskip -0.1in
\end{figure}

\textbf{Manifold Assumption}.\quad The data manifold hypothesis indicates that high-dimensional data often lies in an unknown lower-dimensional manifold embedded in ambient space \cite{doi:10.1126/science.290.5500.2323, DBLP:journals/neco/BelkinN03, DBLP:conf/nips/LevinaB04, lunga2013manifold, DBLP:journals/corr/abs-2207-02862} and is supported by strong evidence.
From a theoretical perspective, \citet{DBLP:conf/nips/OzakinG09, DBLP:conf/nips/NarayananM10} prove that when such hypothesis holds, manifold learning and density estimation scale exponentially with the \emph{low intrinsic} dimension, but otherwise scale exponentially with the \emph{high ambient} dimension \cite{cacoullos1964estimation}.
Therefore, as \citet{DBLP:journals/corr/abs-2207-02862} point out, one most plausible explanation for the success of machine learning methods on real-world data is the existence of such lower intrinsic dimension, which enables learning on datasets of fairly reasonable size, which is empirically verified by \citet{pope2021intrinsic}.
Also, for two of the real-world tabular datasets used in this work, we visualize their 3-dimensional manifolds and calculate their intrinsic dimensions in Figure \ref{fig: manifold}.
%
%
%

Our GCDRO algorithm uses an input-weighted graph $G_N$ to approximate the data manifold.
The $k$NN graph is a fundamental and basic way to represent the data structure, and manifold learning is an area with intensive research. 
We have to clarify that manifold learning is not the focus of this paper, which takes the data structure $G_N$ as input to design a DRO objective and optimization algorithm that incorporates data geometric information for more reasonable worst-case distribution.
Notably, our GCDRO achieves significant performance in the experiments even with the simple $k$NN representation of data structure. 
It proves that this direction for geometric-aware DROs is promising, and our proposed method could efficiently leverage the geometric properties encoded in the input graph to mitigate the effects of harmful data points (note that no target information is leaked into $G_N$).
Actually, our GCDRO is compatible with any manifold learning or graph learning method. We do believe that a more accurate estimated data structure with advanced manifold learning algorithms will further boost the performance of GCDRO, and we leave this to future work.

\begin{figure}[h]
    \centering
    \includegraphics[width=0.5\linewidth]{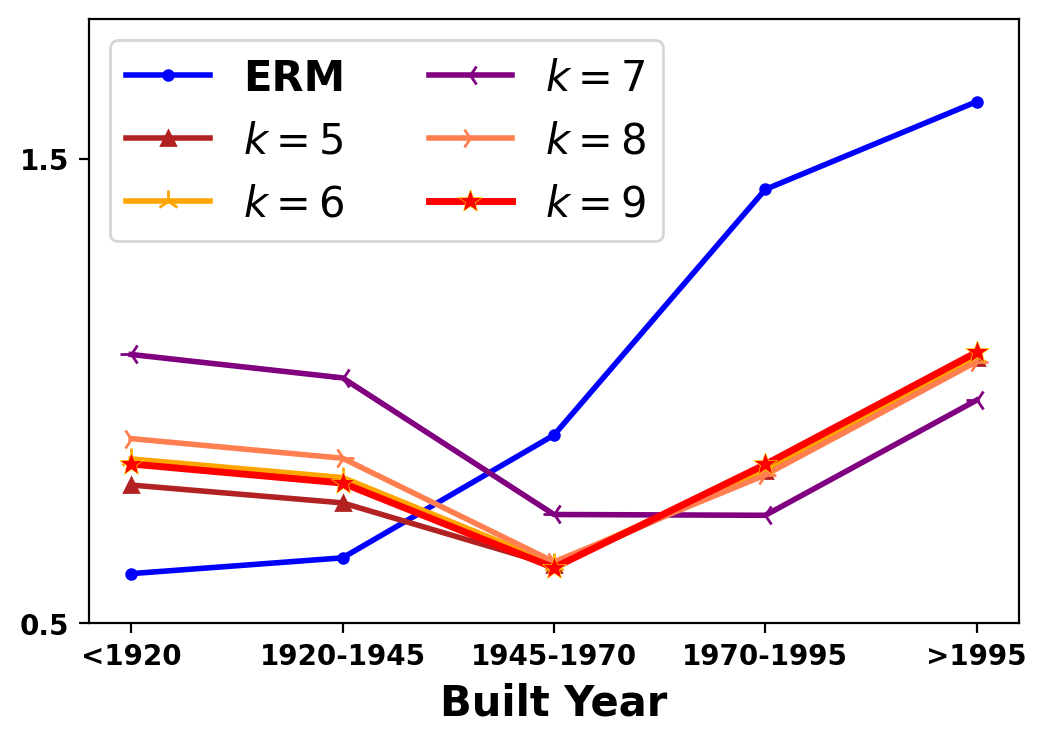}
    \caption{Results with varying $k$.}
    \label{fig:k}
\end{figure}

\textbf{Not Sensitive to $k$}.\quad For the house pricing dataset, we plot the results of our GCDRO with varying $k$s in Figure \ref{fig:k}.
We could see that the performance of our algorithm is not affected much.

\section{Experimental Details}
\textbf{Model \& Loss function.}\quad For simulation data, we use linear models for all methods. For real-world data, we use two-layer MLPs for all methods. 

\textbf{Optimizer.}\quad For all experiments, we use Adam with a learning rate of $1e-3$ in PyTorch for all methods.

\textbf{Hyper-parameters.}\quad For KLDRO, WDRO and $\chi^2$-DRO, we grid search the radius of the uncertainty set within the range of $[1e-3, 2e2]$, and we select the best hyper-parameters according to their testing performances.
For GDRO, we grid search the number of gradient flow steps within the range of $[1e2,2e3]$, the parameter $\beta\in [1,20]$ and we select the best hyper-parameters according to its testing performances.
For DORO, we set the noisy ratio to the ground-truth value for the simulation data, and we grid search the ratio of noisy points within the range of $[1e-2,5e-1]$ for the real-world data.
For HRDRO, we use $L_1$ loss as proposed in \citep{bennouna2022holistic} and grid search $\epsilon\in [1e-3, 1]$.
For GCDRO, we grid search the number of gradient flow steps within the range of $[1e2,2e3]$, $\beta\in[1,20]$ and $\alpha\in [1e-1,1e1]$.
We select the best hyper-parameters according to their testing performances.

Note that in our experiments, we found that model selection without domain information in the validation set is very hard, which is also verified by \cite{zhai2021doro, DBLP:conf/iclr/GulrajaniL21}.
And we believe this is still an open problem and is fairly non-trivial.

\section{Examples on Label Noise}
\begin{theorem}
Assume that the training data is a mixture of $n_c$ clean samples $\{ x_c^{(i)},y_c^{(i)} \}$ drawn from distribution $P_c(X,Y)$ and $n_o$ noisy samples $\{ x_o^{(i)}, y_o^{(i)}  \}$ drawn from distribution $P_o(X,Y)$. Consider a linear data generation process, i.e. $Y=kX+\xi$ and $\xi\perp X, \mathbb{E}[\xi]=0$ and $\mathbb{E}[\xi^2]$ is finite (abbr. $\sigma^2$). The regression model is parameterized as $f(x) = \theta \cdot x$ and trained with Weighted Least Square estimation:
\begin{align}
    &\hat\theta = \arg\min_\theta \sum_{i=1}^{n_c} w_c^{(i)}\| (y_c^{(i)}-\theta\cdot x_c^{(i)})\|^2 + \sum_{i=1}^{n_o} w_o^{(i)}\| (y_o^{(i)}-\theta\cdot x_o^{(i)})\|^2. \\
    s.t. \;\;  &\sum_{i=1}^{n_c} w_c^{(i)} + \sum_{i=1}^{n_o} w_o^{(i)} = 1,
\end{align}
where $w_c^{(i)},w_o^{(i)}\geq 0$ are weights on clean and noisy samples respectively, and $\sigma_c^2<\sigma_o^2$.
Then the variance of the least square estimate $\hat\theta$ is given by:
\begin{align}
    \text{Var}[\hat\theta | X_D] =  \frac{\sum_{i=1}^{n_c} (w_c^{(i)})^2 (x_c^{(i)})^2 \sigma_c^2 + \sum_{i=1}^{n_o} (w_o^{(i)})^2 (x_o^{(i)})^2 \sigma_o^2}{\left[ {\sum_{i=1}^{n_c} w_c^{(i)} (x_c^{(i)})^2  + \sum_{i=1}^{n_o} w_o^{(i)} (x_o^{(i)})^2} \right]^2},
\end{align}
where $X_D = \{ x_c^{(i)} \}\cup \{ x_o^{(i)} \}$ is the sampled covariates in the dataset. Further, the variance of the estimator $\hat\theta$ achieves the minimum if and only if:
\begin{align}
    \forall 1\leq i \leq n_c , 1\leq j\leq n_o, \;\; \gamma(i,j) = w_o^{(j)}/w_c^{(i)} = \sigma_c^2 / \sigma_o^2,
\end{align}
where $\gamma(i,j)$ denotes the sample weight ratio between $i$ and $j$.
\end{theorem}
The theorem is a direct corollary of the following lemma.
\begin{lemma}
Assume that the training data contains $n$ samples $\{ x^{(i)},y^{(i)} \}$. Consider a linear data generation process with heterogeneous noise, i.e. $y^{(i)}=kx^{(i)}+\xi_i$ with $\xi_i\perp X, \mathbb{E}[\xi_i]=0$, and $\mathbb{E}[\xi_i^2]$ is finite. The regression model is parameterized as $f(x) = \theta \cdot x$ and trained with Weighted Least Square estimation:
\begin{align}
    &\hat\theta = \arg\min_\theta \sum_{i=1}^{n} w^{(i)}\| (y^{(i)}-\theta\cdot x^{(i)})\|^2. \\
    s.t. \;\;  &\sum_{i=1}^{n} w^{(i)} = 1,
\end{align}
where $w^{(i)}\geq 0$ are sample weights. Then the variance of the least square estimate $\hat\theta$ is given by:
\begin{align}
    \label{eq:variance_linear}
    \text{Var}[\hat\theta | X_D] =  \frac{\sum_{i=1}^{n} (w^{(i)})^2 (x^{(i)})^2 \sigma_i^2 }{\left[ \sum_{i=1}^{n} w^{(i)} (x^{(i)})^2 \right]^2},
\end{align}
where $X_D = \{ x^{(i)} \}$ is the sampled covariates in the dataset. Further, the variance of the estimator $\hat\theta$ achieves the minimum if and only if:
\begin{align}
    \forall 1\leq i \leq n , 1\leq j\leq n, \;\; w^{(i)}\sigma_i^2 = w^{(j)}\sigma_j^2.
\end{align}
\end{lemma}
\begin{proof}
According to the heterogeneous noise distribution, let $y^{(i)} = x^{(i)} + \epsilon_i$, where $\epsilon_i \sim \mathcal N(0,\sigma_i^2)$.
The least square estimation of $\hat \theta$ is given by:
\begin{align}
    \hat \theta = k + \frac{\sum_{i=1}^{n} w^{(i)} x^{(i)} \epsilon_i}{\sum_{i=1}^{n} w^{(i)} (x^{(i)})^2}.
\end{align}
Since $\mathbb E[\hat\theta|X_D] = k$, we have
\begin{align}
    \text{Var}[\hat\theta | X_D] &= \mathbb E\left|  \frac{\sum_{i=1}^{n} w^{(i)} x^{(i)} \epsilon_i}{\sum_{i=1}^{n} w^{(i)} (x^{(i)})^2} 
    \right|^2 \\
    &= \frac{\sum_{i=1}^{n} (w^{(i)})^2 (x^{(i)})^2 \sigma_i^2 }{\left[ \sum_{i=1}^{n} w^{(i)} (x^{(i)})^2 \right]^2}.
\end{align}
    Next, we solve the minimum of Eq.\ref{eq:variance_linear} w.r.t. sample weights $w^{(i)}$. Let $\alpha_i = w^{(i)}(x^{(i)})^2$. We could formulate the variance in Eq.\ref{eq:variance_linear} as a function of $\alpha = (\alpha_1,...,\alpha_n)$:
\begin{align}
    V(\alpha) = \frac{\sum_{i=1}^{n} \alpha_i^2 \sigma_i^2/(x^{(i)})^2}{\left(\sum_{i=1}^{n} \alpha_i\right)^2}.
\end{align}
Since $V(\lambda\alpha) = V(\alpha)$ for any $\lambda>0$, we could assume $\sum_{i=1}^{n} \alpha_i=1$ without loss of generality. Then the minimization of $V(\alpha)$ is equivalent to:
\begin{align}
    &\min_\alpha V(\alpha) = \sum_{i=1}^{n} \alpha_i^2 \sigma_i^2/(x^{(i)})^2. \\
    s.t.\;\;& \sum_{i=1}^{n} \alpha_i=1.
\end{align}
The first-order KKT condition gives:
\begin{align}
    \exists C, \forall 1\leq i\leq n, \;\;\alpha_i^* = C (x^{(i)})^2 /\sigma_i^2,
\end{align}
from which we can solve:
\begin{align}
\label{eq:solution_weight_min}
    \alpha_i^* = \frac{(x^{(i)})^2/\sigma_i^2} {\sum_{j=1}^n (x^{(j)})^2/\sigma_j^2 }.
\end{align}
Since $\nabla_\alpha^2 V(\alpha) = diag\left[2\sigma_1^2/(x^{(1)})^2,..., 2\sigma_n^2/(x^{(n)})^2\right]$ is always positive definite, Eq.\ref{eq:solution_weight_min} minimizes $V(\alpha)$. Correspondingly $w^{(i)} \propto 1/\sigma_i^2$, which finishes the proof.
\end{proof}

\section{Proofs}

\subsection{Proof of Proposition 2.1}
\begin{proof}
	(1) For KL-divergence as the distance function, we have the following optimization problem under finite samples. 
\begin{equation}
\label{equ:minimax-kl}
	\min_{\theta\in\Theta, \lambda\geq 0}\sup\limits_{\textbf{p}\in\Delta_n} \left\{ \sum_{i=1}^np_i\ell(f_\theta(x_i),y_i)-\lambda\sum_{i=1}^n p_i\log p_i+\lambda(\epsilon-\log n) \right\},
\end{equation}
Solve the inner supremum problem, and the worst-case distribution is like:
\begin{equation}
	p_i = \exp\left(\frac{\ell_i-\eta}{\lambda}-1\right),\ \eta(\ell)=\lambda\log\lambda +\lambda\log\left(\sum_{i=1}^n\exp(\frac{\ell_i}{\lambda}-1)\right),
\end{equation}
and the objective function becomes:
\begin{equation}
	\min_{\theta\in\Theta, \lambda\geq 0} \lambda \log \left(\sum_{i=1}^n \exp(\frac{\ell(f_\theta(x_i),y_i))}{\lambda})\right)+\lambda(\epsilon+\log\lambda -\log n).
\end{equation}
And we could compare the sample weights of different samples as:
\begin{equation}
	\frac{p_i}{p_j} = \exp(\frac{\ell_i-\ell_j}{\lambda}).
\end{equation}

(2) For $\chi^2$-divergence which is defined as $f(x)=(x-1)^2$, we have the following optimization problem.
\begin{equation}
\label{equ:minimax-x2}
	\min_{\theta\in\Theta, \lambda\geq 0}\sup\limits_{\textbf{p}\in\Delta_n} \left\{ \sum_{i=1}^np_i\ell(f_\theta(x_i),y_i) +\lambda\epsilon - \frac{\lambda}{n} \sum_{i=1}^n (np_i-1)^2 \right\}.
\end{equation}
Solve the inner supremum problem, and we have the worst-case distribution like:
\begin{equation}
	p_i = \frac{1}{\lambda n}(\ell_i + \lambda - \eta)_+,
\end{equation}
and the objective function becomes:
\begin{equation}
	\min_{\theta\in\Theta, \lambda\geq 0, \eta\in\mathbb R}\sum_{i=1}^n \frac{1}{2\lambda}(\ell_i + \lambda - \eta)_+^2+\lambda\epsilon + \eta - \frac{\lambda}{2}.
\end{equation}
And we could compare the sample weights of different samples as:
\begin{equation}
	\frac{p_i}{p_j}=\frac{(\ell_i+\lambda-\eta)_+}{(\ell_j+\lambda-\eta)_+},
\end{equation}
if $p_j > 0$. 

(3) For Maximal Mean Discrepancy (MMD) distance, we have the following optimization problem:
\begin{align}
	&\sup\limits_{\textbf{p}}\left\{\sum_{i=1}^n p_i\ell_i +\lambda\epsilon - \lambda (\bf{p}-\frac{1}{n})^TK(\bf{p}-\frac{1}{n})\right\}\\
	\text{s.t.\quad}& \sum_{i=1}^np_i = 1\\
	& p_i \geq 0, \text{for }i = 1, \dots, n
\end{align}
Solve the inner supremum problem, and we have the worst-case distribution like:
\begin{equation}
	p^* = \frac{1}{2\lambda}K^{-1}(\ell-\eta+\frac{2\lambda}{n}K\textbf{1})_+,
\end{equation}
and the objective function becomes:
\begin{equation}
	\min_{\theta\in\Theta,\lambda\geq 0, \eta\in\mathbb R}\frac{1}{4\lambda}(\ell+\frac{2\lambda}{n}K\textbf{1}-\eta)_+K^{-1}(\ell+\frac{2\lambda}{n}K\textbf{1}-\eta)_+ + \lambda\epsilon + \eta - \frac{\lambda}{n^2}\textbf{1}^TK\textbf{1}.
\end{equation}
\end{proof}

\subsection{Proof of Proposition 3.1}
\begin{proof}
	Please refer to the proof of Proposition \ref{proposition: worst-case distribution} for the proof of KL-DRO, $\chi^2$-DRO and MMD DRO.
	For marginal DRO, it is easy to prove following \cite{duchi2022distributionally}.
	For GDRO, it is easy to prove following \cite{liudistributionally}.
\end{proof}

\subsection{Proof of Proposition 3.2}
\label{appendix-3-5}
\begin{proof}
	The proof is based on the Theorem 5 in \cite{chow2017entropy}.
From \cite{chow2017entropy}, we have
\begin{equation}
	\mathcal{R}_N(q^\infty) - \mathcal{R}_N(q(t))\leq e^{-Ct}(\mathcal{R}_N(q^\infty)-\mathcal{R}_N(q^0)).
\end{equation}
Furthermore,
\begin{equation}
	C:=2m\lambda_{\text{sec}}(\hat{L})\lambda_{\text{min}}(\nabla^2\mathcal R_N)\frac{1}{(r+1)^2} > 0,
\end{equation}
and
\begin{equation}
	r = \sqrt{2}k\max_{(i,j)\in E}w_{ij}\frac{\|\text{Hess}\mathcal R_N\|_1}{\lambda_{\text{min}}(\text{Hess}\mathcal R_N)^{1.5}}\frac{1-m}{m^2}\frac{\lambda_{\text{max}}(\hat{L})}{\lambda_{\text{sec}}(\hat{L})^2}\sqrt{\mathcal R_N(q^0)-\mathcal R_N(q^\infty)},
\end{equation}
where $k$ denotes the number of neighbors in the $k$NN graph, $\hat{L}$ is the graph Laplacian matrix, $\lambda_{\text{sec}}, \lambda_{\text{min}}$ are the second smallest and smallest eigenvalue, and
\begin{equation}
	\|\text{Hess}\mathcal R_N\|_1 = \sup_{q\in\mathscr P(G_N)}\|\text{Hess}\mathcal R_N(q)\|_1, \quad\lambda_{\text{min}}(\text{Hess}\mathcal R_N)=\min_{q\in\mathscr{P}(G_N)}\lambda_{\text{min}}(\text{Hess}\mathcal R_N(q)),
\end{equation}
and
\begin{equation}
	m = \frac{1}{2}(\frac{1}{(1+2M)^{\frac{1}{\beta}}})^{N-2}\min\{\frac{1}{(1+2M)^{\frac{1}{\beta}}}), \frac{1}{N}\}.
\end{equation}

Then denote the real worst-case distribution within the $\epsilon(\theta)$-radius discrete Geometric Wasserstein-ball as $q^*$, that is,
	\begin{small}
	\begin{equation}
	\label{equ:p-star}
		q^*
		= \arg\sup\limits_{q:\mathcal{GW}^2_{G_N}(\hat{P}_{tr},q)\leq \epsilon(\theta)}\mathcal R_N(\theta,q),
	\end{equation}	
	\end{small}
and we have
\begin{equation}
	\mathcal{R}_N(q^\infty) -\mathcal{R}_N(q^*)+\mathcal{R}_N(q^*) - \mathcal{R}_N(q(t))\leq e^{-Ct}(\mathcal{R}_N(q^\infty)-\mathcal{R}_N(q^*)+\mathcal{R}_N(q^*)-\mathcal{R}_N(q^0)).
\end{equation}
Therefore, we have
\begin{equation}
	\mathcal{R}_N(q^*)-\mathcal{R}_N(q(t)) \leq e^{-Ct}(\mathcal{R}_N(q^*)-\mathcal{R}_N(q^0))-(1-e^{-Ct})(\mathcal{R}_N(q^\infty)-\mathcal{R}_N(q^*)),
\end{equation}
and
\begin{equation}
	\frac{\mathcal{R}_N(q^*)-\mathcal{R}_N(q(t))}{\mathcal{R}_N(q^*)-\mathcal{R}_N(q^0)}\leq e^{-Ct} - (1-e^{-Ct})\frac{\mathcal{R}_N(q^\infty)-\mathcal{R}_N(q^*)}{\mathcal{R}_N(q^*)-\mathcal{R}_N(q^0)}<e^{-Ct}.
\end{equation}

\end{proof}

\subsection{Proof of Proposition 3.3}
\begin{proof}
	It is easy to prove that the final state of $\mathcal R_N(\theta,q)$ w.r.t. $q$ is given as
	\begin{equation}
		q^{\infty}_i = \frac{1}{Z} \exp(\frac{\ell_i - \alpha(\sum_{j\in N(i)}q^{\infty}_jw_{ij}(\ell_i-\ell_j)^2)}{\beta}),
	\end{equation}
	where 
	\begin{equation}
	\label{equ:Z}
	Z=\sum_{i=1}^N \exp(\frac{\ell_i - \alpha(\sum_{j\in N(i)}q^{\infty}_jw_{ij}(\ell_i-\ell_j)^2)}{\beta}).	
	\end{equation}
	
	(1) When $\beta\rightarrow\infty$, $q_i^\infty \rightarrow \frac{1}{N}$.
	When $\beta \ll \infty$, the gradient flow is like:
	\begin{equation}
		\frac{dq_i}{dt}=\sum_{(i,j)\in E} w_{ij}\xi_{ij}\bigg(\ell_i-\ell_j+\beta(\log q_j-\log q_i)+\alpha\big(\sum_{h\in N(j)}(\ell_h-\ell_j)^2w_{hj}q_h-\sum_{h\in N(i)}(\ell_h-\ell_i)^2w_{hi}q_h \big)\bigg),
	\end{equation}
	and 
	\begin{equation}
	\xi_{ij}(v):= v\cdot\big(\mathbb I(v>0)q_j + \mathbb I(v\leq 0)q_i\big).
	\end{equation}
	Therefore, when $q_i>q_j$ and $\ell_i>\ell_j$, we have $\log q_j-\log q_i<0$, which decreases the gradient of $q_i$.
	Thus, the entropy term prompts the sample weights to be smooth between neighbors.
	When the sample weight of sample $i$ is larger than its neighbors, this term will decrease the gradient of $q_i$ to prevent it from gaining too much weights.\\
	(3) Under the assumptions, we have
	\begin{align}
		&(\sum_{j\in N(i)}q_j^\infty w_{ij}(\ell_i+\Delta_i-\ell_j)^2-\sum_{j\in N(i)}q_j^\infty w_{ij}(\ell_i-\ell_j)^2)\\
		=&\sum_{j\in N(i)}q_j^\infty w_{ij}(2\ell_i-2\ell_j+\Delta_i)\Delta_i\\
		\geq &\Delta_i \big(\sum_{j\in N(i)}q_j^\infty w_{ij}(\Delta_i-2L_x\|x_i-x_j\|_2)\big)\\
		> & \Delta_i.
	\end{align}
	Therefore, define $\delta_i = \ell_i^{\text{noisy}}-\ell_i$, it is easy to prove that for $\alpha>0$
 \begin{align}
     \xi_{\text{GCDRO}}(i,j) < \xi_{\text{GDRO}}(i,j),
 \end{align}
 and when $\alpha>\frac{1}{\sum_{k\in N(i)}q_kw_{ik}(2\ell_i-2\ell_k+\delta_i)}$, we have $\xi_{\text{GCDRO}}(i,j)< 0$.
\end{proof}


%% file: main.bbl
\begin{thebibliography}{41}
\providecommand{\natexlab}[1]{#1}
\providecommand{\url}[1]{\texttt{#1}}
\expandafter\ifx\csname urlstyle\endcsname\relax
  \providecommand{\doi}[1]{doi: #1}\else
  \providecommand{\doi}{doi: \begingroup \urlstyle{rm}\Url}\fi

\bibitem[Agarwal and Zhang(2022)]{agarwal2022minimax}
A.~Agarwal and T.~Zhang.
\newblock Minimax regret optimization for robust machine learning under distribution shift.
\newblock \emph{arXiv preprint arXiv:2202.05436}, 2022.

\bibitem[Belkin and Niyogi(2003)]{DBLP:journals/neco/BelkinN03}
M.~Belkin and P.~Niyogi.
\newblock Laplacian eigenmaps for dimensionality reduction and data representation.
\newblock \emph{Neural Comput.}, 15\penalty0 (6):\penalty0 1373--1396, 2003.
\newblock \doi{10.1162/089976603321780317}.
\newblock URL \url{https://doi.org/10.1162/089976603321780317}.

\bibitem[Bennouna and Van~Parys(2022)]{bennouna2022holistic}
A.~Bennouna and B.~Van~Parys.
\newblock Holistic robust data-driven decisions.
\newblock \emph{arXiv preprint arXiv:2207.09560}, 2022.

\bibitem[Blanchet and Murthy(2019)]{blanchet2019quantifying}
J.~Blanchet and K.~Murthy.
\newblock Quantifying distributional model risk via optimal transport.
\newblock \emph{Mathematics of Operations Research}, 44\penalty0 (2):\penalty0 565--600, 2019.

\bibitem[Blanchet et~al.(2019{\natexlab{a}})Blanchet, Kang, and Murthy]{blanchet2019robust}
J.~Blanchet, Y.~Kang, and K.~Murthy.
\newblock Robust wasserstein profile inference and applications to machine learning.
\newblock \emph{Journal of Applied Probability}, 56\penalty0 (3):\penalty0 830--857, 2019{\natexlab{a}}.

\bibitem[Blanchet et~al.(2019{\natexlab{b}})Blanchet, Kang, Murthy, and Zhang]{DBLP:conf/wsc/BlanchetKMZ19}
J.~H. Blanchet, Y.~Kang, K.~R.~A. Murthy, and F.~Zhang.
\newblock Data-driven optimal transport cost selection for distributionally robust optimization.
\newblock In \emph{2019 Winter Simulation Conference, {WSC} 2019, National Harbor, MD, USA, December 8-11, 2019}, pages 3740--3751. {IEEE}, 2019{\natexlab{b}}.
\newblock \doi{10.1109/WSC40007.2019.9004785}.
\newblock URL \url{https://doi.org/10.1109/WSC40007.2019.9004785}.

\bibitem[Brown et~al.(2022)Brown, Caterini, Ross, Cresswell, and Loaiza{-}Ganem]{DBLP:journals/corr/abs-2207-02862}
B.~C.~A. Brown, A.~L. Caterini, B.~L. Ross, J.~C. Cresswell, and G.~Loaiza{-}Ganem.
\newblock The union of manifolds hypothesis and its implications for deep generative modelling.
\newblock \emph{CoRR}, abs/2207.02862, 2022.
\newblock \doi{10.48550/arXiv.2207.02862}.
\newblock URL \url{https://doi.org/10.48550/arXiv.2207.02862}.

\bibitem[Cacoullos(1964)]{cacoullos1964estimation}
T.~Cacoullos.
\newblock Estimation of a multivariate density.
\newblock Technical report, University of Minnesota, 1964.

\bibitem[Chow et~al.(2017)Chow, Li, and Zhou]{chow2017entropy}
S.-N. Chow, W.~Li, and H.~Zhou.
\newblock Entropy dissipation of fokker-planck equations on graphs.
\newblock \emph{arXiv preprint arXiv:1701.04841}, 2017.

\bibitem[Diakonikolas and Kane(2018)]{diakonikolas2018algorithmic}
I.~Diakonikolas and D.~M. Kane.
\newblock Algorithmic high-dimensional robust statistics.
\newblock \emph{Webpage http://www. iliasdiakonikolas. org/simons-tutorial-robust. html}, 2018.

\bibitem[Diakonikolas et~al.(2022)Diakonikolas, Kane, Pensia, and Pittas]{diakonikolas2022streaming}
I.~Diakonikolas, D.~M. Kane, A.~Pensia, and T.~Pittas.
\newblock Streaming algorithms for high-dimensional robust statistics.
\newblock In \emph{International Conference on Machine Learning}, pages 5061--5117. PMLR, 2022.

\bibitem[Dua and Graff(2017)]{Dua:2019}
D.~Dua and C.~Graff.
\newblock Uci machine learning repository, 2017.
\newblock URL \url{http://archive.ics.uci.edu/ml}.

\bibitem[Duchi et~al.(2022)Duchi, Hashimoto, and Namkoong]{duchi2022distributionally}
J.~Duchi, T.~Hashimoto, and H.~Namkoong.
\newblock Distributionally robust losses for latent covariate mixtures.
\newblock \emph{Operations Research}, 2022.

\bibitem[Duchi and Namkoong(2021)]{duchi2021learning}
J.~C. Duchi and H.~Namkoong.
\newblock Learning models with uniform performance via distributionally robust optimization.
\newblock \emph{The Annals of Statistics}, 49\penalty0 (3):\penalty0 1378--1406, 2021.

\bibitem[Esposito et~al.(2021)Esposito, Patacchini, Schlichting, and Slep{\v{c}}ev]{esposito2021nonlocal}
A.~Esposito, F.~S. Patacchini, A.~Schlichting, and D.~Slep{\v{c}}ev.
\newblock Nonlocal-interaction equation on graphs: gradient flow structure and continuum limit.
\newblock \emph{Archive for Rational Mechanics and Analysis}, 240\penalty0 (2):\penalty0 699--760, 2021.

\bibitem[Friston(2010)]{friston2010free}
K.~Friston.
\newblock The free-energy principle: a unified brain theory?
\newblock \emph{Nature reviews neuroscience}, 11\penalty0 (2):\penalty0 127--138, 2010.

\bibitem[Frogner et~al.(2019)Frogner, Claici, Chien, and Solomon]{frogner2019incorporating}
C.~Frogner, S.~Claici, E.~Chien, and J.~Solomon.
\newblock Incorporating unlabeled data into distributionally robust learning.
\newblock \emph{arXiv preprint arXiv:1912.07729}, 2019.

\bibitem[Fu et~al.(1990)Fu, Shen, Yao, and Hou]{fu1990physical}
X.~Fu, W.~Shen, T.~Yao, and W.~Hou.
\newblock Physical chemistry.
\newblock \emph{Higher Education, Beijing}, 1990.

\bibitem[Gao and Kleywegt(2022)]{gao2022distributionally}
R.~Gao and A.~Kleywegt.
\newblock Distributionally robust stochastic optimization with wasserstein distance.
\newblock \emph{Mathematics of Operations Research}, 2022.

\bibitem[Gao et~al.(2022)Gao, Chen, and Kleywegt]{gao2022wasserstein}
R.~Gao, X.~Chen, and A.~J. Kleywegt.
\newblock Wasserstein distributionally robust optimization and variation regularization.
\newblock \emph{Operations Research}, 2022.

\bibitem[Gulrajani and Lopez{-}Paz(2021)]{DBLP:conf/iclr/GulrajaniL21}
I.~Gulrajani and D.~Lopez{-}Paz.
\newblock In search of lost domain generalization.
\newblock In \emph{9th International Conference on Learning Representations, {ICLR} 2021, Virtual Event, Austria, May 3-7, 2021}. OpenReview.net, 2021.
\newblock URL \url{https://openreview.net/forum?id=lQdXeXDoWtI}.

\bibitem[Hu et~al.(2018)Hu, Niu, Sato, and Sugiyama]{hu2018does}
W.~Hu, G.~Niu, I.~Sato, and M.~Sugiyama.
\newblock Does distributionally robust supervised learning give robust classifiers?
\newblock In \emph{International Conference on Machine Learning}, pages 2029--2037. PMLR, 2018.

\bibitem[Huber(1992)]{huber1992robust}
P.~J. Huber.
\newblock Robust estimation of a location parameter.
\newblock \emph{Breakthroughs in statistics: Methodology and distribution}, pages 492--518, 1992.

\bibitem[Kingma and Ba(2015)]{adam}
D.~P. Kingma and J.~Ba.
\newblock Adam: {A} method for stochastic optimization.
\newblock In Y.~Bengio and Y.~LeCun, editors, \emph{3rd International Conference on Learning Representations, {ICLR} 2015, San Diego, CA, USA, May 7-9, 2015, Conference Track Proceedings}, 2015.
\newblock URL \url{http://arxiv.org/abs/1412.6980}.

\bibitem[Klivans et~al.(2018)Klivans, Kothari, and Meka]{klivans2018efficient}
A.~Klivans, P.~K. Kothari, and R.~Meka.
\newblock Efficient algorithms for outlier-robust regression.
\newblock In \emph{Conference On Learning Theory}, pages 1420--1430. PMLR, 2018.

\bibitem[Levina and Bickel(2004)]{DBLP:conf/nips/LevinaB04}
E.~Levina and P.~J. Bickel.
\newblock Maximum likelihood estimation of intrinsic dimension.
\newblock In \emph{Advances in Neural Information Processing Systems 17 [Neural Information Processing Systems, {NIPS} 2004, December 13-18, 2004, Vancouver, British Columbia, Canada]}, pages 777--784, 2004.
\newblock URL \url{https://proceedings.neurips.cc/paper/2004/hash/74934548253bcab8490ebd74afed7031-Abstract.html}.

\bibitem[Liu et~al.(2022{\natexlab{a}})Liu, Shen, Cui, Zhou, Kuang, and Li]{liu2021distributionally}
J.~Liu, Z.~Shen, P.~Cui, L.~Zhou, K.~Kuang, and B.~Li.
\newblock Distributionally robust learning with stable adversarial training.
\newblock \emph{IEEE TKDE}, 2022{\natexlab{a}}.

\bibitem[Liu et~al.(2022{\natexlab{b}})Liu, Wu, Li, and Cui]{liudistributionally}
J.~Liu, J.~Wu, B.~Li, and P.~Cui.
\newblock Distributionally robust optimization with data geometry.
\newblock In \emph{Advances in Neural Information Processing Systems}, 2022{\natexlab{b}}.

\bibitem[Lunga et~al.(2013)Lunga, Prasad, Crawford, and Ersoy]{lunga2013manifold}
D.~Lunga, S.~Prasad, M.~M. Crawford, and O.~Ersoy.
\newblock Manifold-learning-based feature extraction for classification of hyperspectral data: A review of advances in manifold learning.
\newblock \emph{IEEE Signal Processing Magazine}, 31\penalty0 (1):\penalty0 55--66, 2013.

\bibitem[Namkoong and Duchi(2017)]{namkoong2017variance}
H.~Namkoong and J.~C. Duchi.
\newblock Variance-based regularization with convex objectives.
\newblock \emph{Advances in neural information processing systems}, 30, 2017.

\bibitem[Narayanan and Mitter(2010)]{DBLP:conf/nips/NarayananM10}
H.~Narayanan and S.~K. Mitter.
\newblock Sample complexity of testing the manifold hypothesis.
\newblock In J.~D. Lafferty, C.~K.~I. Williams, J.~Shawe{-}Taylor, R.~S. Zemel, and A.~Culotta, editors, \emph{Advances in Neural Information Processing Systems 23: 24th Annual Conference on Neural Information Processing Systems 2010. Proceedings of a meeting held 6-9 December 2010, Vancouver, British Columbia, Canada}, pages 1786--1794. Curran Associates, Inc., 2010.
\newblock URL \url{https://proceedings.neurips.cc/paper/2010/hash/8a1e808b55fde9455cb3d8857ed88389-Abstract.html}.

\bibitem[Ozakin and Gray(2009)]{DBLP:conf/nips/OzakinG09}
A.~Ozakin and A.~G. Gray.
\newblock Submanifold density estimation.
\newblock In Y.~Bengio, D.~Schuurmans, J.~D. Lafferty, C.~K.~I. Williams, and A.~Culotta, editors, \emph{Advances in Neural Information Processing Systems 22: 23rd Annual Conference on Neural Information Processing Systems 2009. Proceedings of a meeting held 7-10 December 2009, Vancouver, British Columbia, Canada}, pages 1375--1382. Curran Associates, Inc., 2009.
\newblock URL \url{https://proceedings.neurips.cc/paper/2009/hash/2ac2406e835bd49c70469acae337d292-Abstract.html}.

\bibitem[Pope et~al.(2021)Pope, Zhu, Abdelkader, Goldblum, and Goldstein]{pope2021intrinsic}
P.~Pope, C.~Zhu, A.~Abdelkader, M.~Goldblum, and T.~Goldstein.
\newblock The intrinsic dimension of images and its impact on learning.
\newblock \emph{ICLR}, 2021.

\bibitem[Reichl(1999)]{reichl1999modern}
L.~E. Reichl.
\newblock A modern course in statistical physics, 1999.

\bibitem[Roweis and Saul(2000)]{doi:10.1126/science.290.5500.2323}
S.~T. Roweis and L.~K. Saul.
\newblock Nonlinear dimensionality reduction by locally linear embedding.
\newblock \emph{Science}, 290\penalty0 (5500):\penalty0 2323--2326, 2000.
\newblock \doi{10.1126/science.290.5500.2323}.
\newblock URL \url{https://www.science.org/doi/abs/10.1126/science.290.5500.2323}.

\bibitem[Sinha et~al.(2018)Sinha, Namkoong, and Duchi]{sinha2017certifying}
A.~Sinha, H.~Namkoong, and J.~C. Duchi.
\newblock Certifying some distributional robustness with principled adversarial training.
\newblock In \emph{6th International Conference on Learning Representations, {ICLR} 2018, Vancouver, BC, Canada, April 30 - May 3, 2018, Conference Track Proceedings}. OpenReview.net, 2018.
\newblock URL \url{https://openreview.net/forum?id=Hk6kPgZA-}.

\bibitem[S{\l}owik and Bottou(2022)]{slowik2022distributionally}
A.~S{\l}owik and L.~Bottou.
\newblock On distributionally robust optimization and data rebalancing.
\newblock In \emph{International Conference on Artificial Intelligence and Statistics}, pages 1283--1297. PMLR, 2022.

\bibitem[Staib and Jegelka(2019)]{staib2019distributionally}
M.~Staib and S.~Jegelka.
\newblock Distributionally robust optimization and generalization in kernel methods.
\newblock \emph{Advances in Neural Information Processing Systems}, 32, 2019.

\bibitem[Villani(2021)]{optimaltransport}
C.~Villani.
\newblock Topics in optimal transportation.
\newblock 58, 2021.

\bibitem[Wang et~al.(2019)Wang, Yu, Zheng, Gan, Gai, Ye, Li, Zhou, Huang, Ma, Huang, Guo, Zhang, Lin, Zhao, Li, Smola, and Zhang]{DBLP:journals/corr/abs-1909-01315}
M.~Wang, L.~Yu, D.~Zheng, Q.~Gan, Y.~Gai, Z.~Ye, M.~Li, J.~Zhou, Q.~Huang, C.~Ma, Z.~Huang, Q.~Guo, H.~Zhang, H.~Lin, J.~Zhao, J.~Li, A.~J. Smola, and Z.~Zhang.
\newblock Deep graph library: Towards efficient and scalable deep learning on graphs.
\newblock \emph{CoRR}, abs/1909.01315, 2019.
\newblock URL \url{http://arxiv.org/abs/1909.01315}.

\bibitem[Zhai et~al.(2021)Zhai, Dan, Kolter, and Ravikumar]{zhai2021doro}
R.~Zhai, C.~Dan, Z.~Kolter, and P.~Ravikumar.
\newblock Doro: Distributional and outlier robust optimization.
\newblock In \emph{International Conference on Machine Learning}, pages 12345--12355. PMLR, 2021.

\end{thebibliography}
